\documentclass{article}

\usepackage{microtype}
\usepackage{graphicx}
\usepackage{booktabs} 
\usepackage{multirow}
\usepackage{hyperref}

\usepackage{caption}
\usepackage{subcaption}
\usepackage[table]{xcolor}
\newcommand{\mycc}{\cellcolor{lightgray}}

\newcommand{\xxnote}[3]{}
\renewcommand{\xxnote}[3]{\color{#2}{#1: #3}}

\newcommand{\LP}[1]{}
\newcommand{\LPr}[1]{}
\newcommand{\RB}[1]{}
\newcommand{\venky}[1]{}
\newcommand{\chenyu}[1]{}

\newcommand{\method}{HiSS}

\newcommand{\csp}{CSP}

\newcommand{\benchmark}{CSP-Bench}

\usepackage{makecell}
\usepackage[accepted]{icml2024}

\usepackage{amsmath}
\usepackage{amssymb}
\usepackage{mathtools}
\usepackage{amsthm}
\usepackage{enumitem}

\DeclareMathOperator*{\argmin}{arg\,min}

\usepackage[capitalize,noabbrev]{cleveref}

\theoremstyle{plain}

\theoremstyle{definition}

\theoremstyle{remark}

\usepackage[textsize=tiny]{todonotes}

\usepackage{tikz}
\usetikzlibrary{calc, shapes.geometric, arrows}
\tikzstyle{state} = [circle, text centered, draw=black, fill=red!30]
\tikzstyle{model} = [rectangle, minimum width=3cm, minimum height=1cm, text centered, draw=black, fill=orange!30]
\tikzstyle{arrow} = [thick,->,>=stealth]
\tikzstyle{backarrow} = [thick,<-,>=stealth]

\icmltitlerunning{\method{}: Hierarchical State Space Models}

\begin{document}
\twocolumn[
\icmltitle{Hierarchical State Space Models \\for Continuous Sequence-to-Sequence Modeling}





\icmlsetsymbol{equal}{*}

\begin{icmlauthorlist}
\icmlauthor{Raunaq Bhirangi}{cmu,fair}
\icmlauthor{Chenyu Wang}{nyu}
\icmlauthor{Venkatesh Pattabiraman}{nyu}
\icmlauthor{Carmel Majidi}{cmu}
\icmlauthor{Abhinav Gupta}{cmu}
\icmlauthor{Tess Hellebrekers}{fair}
\icmlauthor{Lerrel Pinto}{nyu}
\end{icmlauthorlist}

\icmlaffiliation{cmu}{Carnegie Mellon University, Pittsburgh, USA}
\icmlaffiliation{fair}{FAIR, Meta}
\icmlaffiliation{nyu}{New York University, NYC, USA}

\icmlcorrespondingauthor{Raunaq Bhirangi}{rbhirang@cs.cmu.edu}

\icmlkeywords{Machine Learning, ICML}

\vskip 0.3in
]


\printAffiliationsAndNotice{}  

\begin{abstract}

Reasoning from sequences of raw sensory data is a ubiquitous problem across fields ranging from medical devices to robotics. These problems often involve using long sequences of raw sensor data (e.g. magnetometers, piezoresistors) to predict sequences of desirable physical quantities (e.g. force, inertial measurements). While classical approaches are powerful for locally-linear prediction problems, they often fall short when using real-world sensors. These sensors are typically non-linear, are affected by extraneous variables (e.g. vibration), and exhibit data-dependent drift. For many problems, the prediction task is exacerbated by small labeled datasets since obtaining ground-truth labels requires expensive equipment. In this work, we present Hierarchical State-Space models (\method{}), a conceptually simple, new technique for continuous sequential prediction. \method{} stacks structured state-space models on top of each other to create a temporal hierarchy. Across six real-world sensor datasets, from tactile-based state prediction to accelerometer-based inertial measurement, \method{} outperforms state-of-the-art sequence models such as causal Transformers, LSTMs, S4, and Mamba by at least $23\%$ on MSE. Our experiments further indicate that \method{} demonstrates efficient scaling to smaller datasets and is compatible with existing data-filtering techniques. Code, datasets and videos can be found on~\url{https://hiss-csp.github.io} 

\end{abstract}

\section{Introduction}
\label{sec:introduction}

Sensors are ubiquitous. From air conditioners to smartphones, automated systems analyze sensory data sequences to control various parameters. This class of problems - continuous sequence-to-sequence prediction from streaming sensory data - is central to real-time decision-making and control~\cite{schutze2004real, stetco2019machine}. Yet, it has received limited attention compared to discrete sequence problems in domains like language~\cite{devlin2018bert} and computer vision~\cite{deng2009imagenet}.

\begin{figure}[t]
    \centering
    \includegraphics[width=\linewidth]{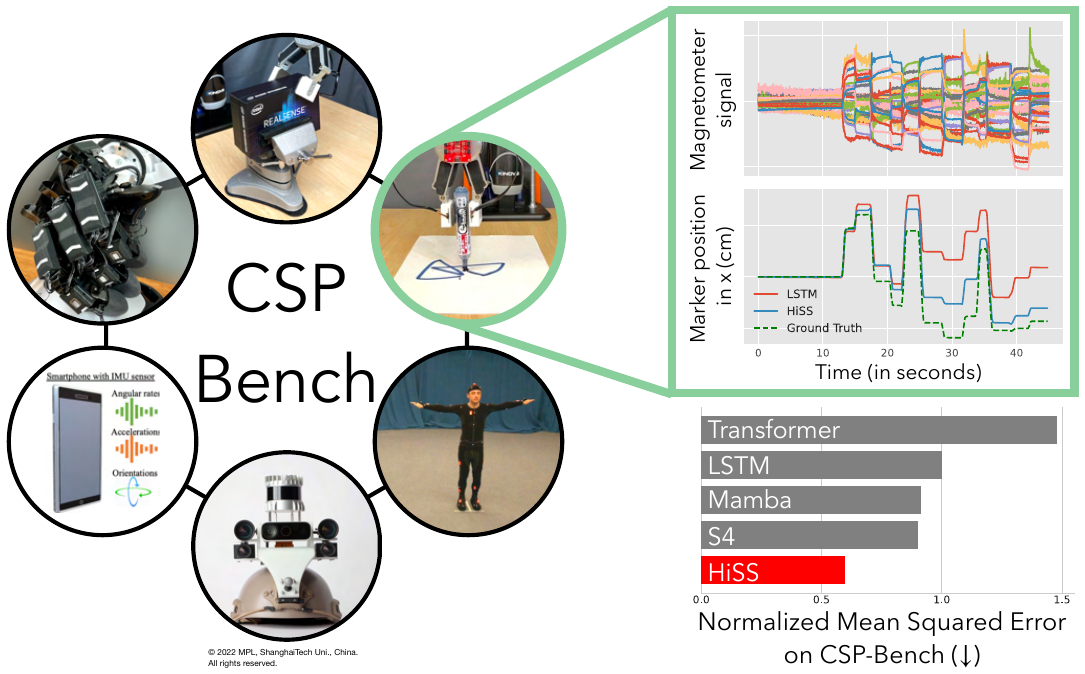}
    \caption{\benchmark{} is a publicly accessible benchmark for continuous sequence prediction on real-world sensory data. We show that Hierarchical State Space Models (\method{}) improve over conventional sequence models on sequential sensory prediction tasks.}
    \label{fig:fig1}
    \vspace{-0.3in}
\end{figure}
Existing approaches for prediction from sensory data have largely relied on model-based solutions~\cite{welch1995introduction, daum2005nonlinear}. However, these approaches require domain expertise and accurate modeling of complex system dynamics, which is often intractable in real-world applications. Moreover, sensory data contains noise and sensor-specific drift that must be accounted for to achieve high predictive performance~\cite{liu2020tlio}. In this work, we investigate deep sequence-to-sequence models that can address these challenges by learning directly from raw sensory streams.

However, to make progress on continuous sequence prediction (CSP), we first need a representative benchmark to measure performance. Most prior works in CSP focus on a single class of sensors~\cite{herath2020ronin, liu2020tlio}, making it difficult to develop general-purpose algorithms. To address this, we created \benchmark{}, a benchmark consisting of six real-world labeled datasets. This collection consists of three datasets created in-house and three curated from prior work -- a cumulative 40 hours of real-world data.

Given data from \benchmark{}, an obvious modeling choice is to use state-of-the-art sequence models like LSTMs or Transformers. However, sensory data is high-frequency, leading to long sequences of highly correlated data. For such data, Transformers quickly run out of memory, as they scale quadratically in complexity with sequence length~\cite{vaswani2017attention}, while LSTMs require significantly larger hidden states~\cite{kuchaiev2017factorization}. 
Deep State Space Models (SSMs)~\cite{gu2021efficiently,gu2023mamba} are a promising new class of sequence models. These models have been shown to effectively handle long context lengths while scaling linearly with sequence length in time and memory complexity, with strong results on audio~\cite{goel2022s} and language modeling. On \benchmark{}, we find that SSMs consistently outperform LSTMs and Transformers with an average of 10\% improvement on MSE metrics (see Section~\ref{sec:experiments}). But can we do better?

A key insight into continuous sensor data is that it has a significant amount of temporal structure and redundancies. While SSMs are powerful for modeling this type of data, they are still temporally flat in nature, i.e. every sample in the sequence is reasoned with every other sample. Therefore, inspired by work in hierarchical modeling~\cite{you2019hierarchical, thu2021hihar}, we propose Hierarchical State-Space Models (\method{}). \method{} stacks two SSMs with different temporal resolutions on top of each other. The lower-level SSM temporally chunks the larger full-sequence data into smaller sequences and outputs local features, while the higher-level SSM operates on the smaller sequence of local features to output global sequence prediction. This leads to further improved performance on \benchmark{}, outperforming the best flat SSMs by 23\% median MSE performance across tasks. We summarize the contributions of this paper as follows: 

\begin{enumerate}[leftmargin=*]
    \item We release \benchmark{}, the largest publicly accessible benchmark for continuous sequence-to-sequence prediction for multiple sensor datasets. (Section \ref{sec:datasets})
    \item We show that SSMs outperform prior SOTA models like LSTMs and Transformers on \benchmark{}. (Section \ref{sec:ssm-v-flat})
    \item We propose \method{}, a hierarchical sequence modeling architecture that \textit{further} improves upon SSMs across tasks in \benchmark{}. (Section \ref{sec:approach})
    \item We show that \method{} increases sample efficiency with smaller datasets, and is compatible with standard sensor pre-processing techniques such as low-pass filtering. (Sections \ref{sec:lowpass}, \ref{sec:small-dsets})
\end{enumerate}

\section{Related Work}
\subsection{Sequence-to-sequence prediction for sensory data}
Most real world control systems, such as wind turbine condition monitoring~\cite{stetco2019machine}, MRI recognition~\cite{kong2016recognizing} and inertial odometry~\cite{amini2011accelerometer,TILO}, often process noisy sensory data to deduce environmental states. Traditionally, these problems were solved as estimation and control problems using filtering techniques, like the Kalman Filter~\cite{mathieu2012state, simon2006optimal}, that still require complex sensor models. Deep learning has shown promise in domains without analytical models, yet many solutions continue to be sensor-specific~\cite{yan2018ridi, herath2020ronin}.

More recently, a number of works~\cite{hasani2022closed,ma2022mega,rusch2021long,morrill2021neural, orvieto2023resurrecting} have been directed at developing neural architectures that improve over conventional sequence models in modeling long-range dependencies. This bodes well for learning sensory prediction models which must naturally reason over long sequences owing to the high frequency nature of sensory data. To the best of our knowledge, however, none of these models have been evaluated on continuous sensing data beyond audio~\cite{goel2022s}. In this work, we focus on deep state space models (SSMs)~\cite{gu2021efficiently, poli2023hyena, smith2022simplified, gu2023mamba, hasani2022closed}, an emerging class of models in long range neural sequence modeling. We benchmark deep SSMs on six sequence-to-sequence prediction tasks on sensors like ReSkin, XELA, accelerometers, and gyroscopes.

\subsection{Hierarchical Modeling}
Incorporating temporal hierarchies into sequence modeling architectures has been shown to improve performance across a number of tasks like sleep classification~\cite{wang2023s4sleep}, recommender systems~\cite{you2019hierarchical}, human activity recognition~\cite{thu2021hihar} and reinforcement learning~\cite{sutton1999between, gardiol2000hierarchical, kulkarni2016hierarchical}. \method{} is inspired by this line of work and extends it to SSMs for continuous seq-to-seq tasks.

\subsection{Data for Continuous Sequence Prediction}
A primary challenge with developing general models for continuous sequence prediction is the lack of a concrete evaluation benchmark. Odometry/SLAM datasets~\cite{geiger2013vision, maddern20171} are viable candidates~\cite{chang2019argoverse, sun2020scalability} for CSP datasets. But most data across sensory modalities like audio~\cite{warden2018speech, gemmeke2017audio}, ECG~\cite{moody2001impact, wagner2020ptb}, IMU~\cite{chavarriaga2013opportunity, micucci2017unimib, chen2021deep} and tactile sensing~\cite{pinto2016curious,funabashi2019morphology, bhirangi2023all} is labeled sparsely only at the sequence level. 

The recent proliferation of sensors in smartphones and other smart devices has resulted in renewed interest in creating labeled datasets for CSP~\cite{chen2018oxiod, herath2020ronin}. A common setting is to use a motion capture system to obtain dense, sequential labels for sensory data from inexpensive IMU sensors~\cite{trumble2017total, gao2022vector}. In this work, we curate three such datasets as part of \benchmark{}: a continuous sequence prediction benchmark.

Another category of sensors of significant interest for CSP are touch sensors. Touch sensors capture the dynamics of contact between the robot and its surroundings. Deep learning and rapid prototyping have driven a rapid surge across a range of tactile modalities from optical~\cite{lambeta2020digit, yuan2017gelsight} to capacitative~\cite{sonar2018any} and magnetic sensing~\cite{tomo2018new,bhirangi2021reskin}. Most work on continuously reasoning over tactile data is directed towards policy learning~\cite{guzey2023see, guzey2023dexterity,  calandra2018more}, where small datasets and confounding factors make it difficult to evaluate the efficacy of architectures for CSP. In this work, we set up supervised learning problems to investigate sequence-to-sequence models for two magnetic tactile sensors: ReSkin~\cite{bhirangi2021reskin} and XELA~\cite{tomo2018new}.

\section{Background}
\label{sec:background}

\subsection{Sequence-to-sequence Prediction}
\label{sec:seq-to-seq}
\begin{figure}[htbp]
    \centering
    \includegraphics[width=0.8\linewidth]{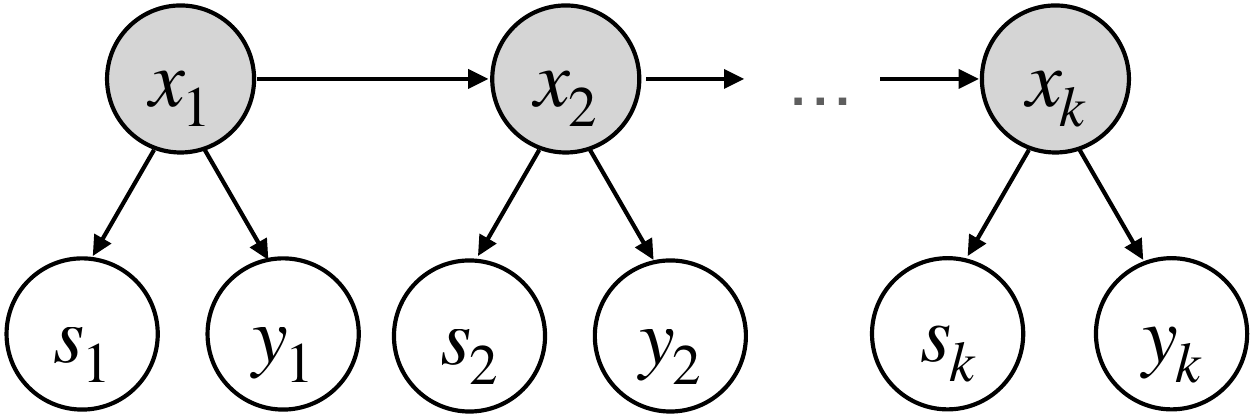}
    \caption{Hidden Markov Model for a two-sensor system. $X$ is a data-generating process. Sensor, $S$, and output, $Y$, are two observable processes.}
    \label{fig:hmm}
\end{figure}

Consider a data-generating process described by the Hidden Markov Model in Figure \ref{fig:hmm}. The observable processes -- sensor, $S$, and output, $Y$, represent two measurement devices that capture the evolution of the unobserved latent process, $X$. Generally, $S$ is a noisy, low-cost device like an accelerometer, and $Y$ is a precise, expensive labeling system like Motion Capture. The goal is to learn a model that allows us to estimate $Y$ using data sequences from $S$.

The CSP problem involves estimating the probability of the $t$-th output observation, $y_t$, given the history of input observations, $s_{1:t}$. For the experiments listed in this paper, we approximate this probability by a Gaussian with constant standard deviation, ie. $p(y_t | s_1, \dots s_t) = \mathcal{N}(\mu_\theta (s_{1:t}), \sigma^2 I)$, where $\sigma$ is a constant, and parameterize $\mu_\theta$ by a deep sequence model. Our goal is to find the maximum likelihood estimator for this distribution -- $\argmin_\theta \sum_t \| y_t - \mu_\theta (s_{1:t}) \|^2$. Therefore, our models are trained to minimize MSE loss over the length of the output sequence.

\subsection{Deep State Space Models}
Deep State Space Models (SSMs) build on simple state space models for sequence-to-sequence modeling. In its general form, a linear state space model may be written as,
\begin{equation*}
\begin{aligned}
    x'(t) &= \mathbf{A}(t)x(t) + \mathbf{B}(t)u(t) \\
    y(t) &= \mathbf{C}(t)x(t) + \mathbf{D}(t)u(t),
\end{aligned}
    \label{eq:ssm}
\end{equation*}
mapping a 1-D input sequence $u(t) \in \mathbb{R}$ to a 1-D output sequence $y(t) \in \mathbb{R}$ through an implicit N-D latent state sequence $x(t) \in \mathbb{R}^n$. Concretely, deep SSMs seek to use stacks of this simple model in a neural sequence modeling architecture, where the parameters, $\mathbf{A}, \textbf{B}, \mathbf{C}$ and $\mathbf{D}$ for each layer can be learned via gradient descent.

SSMs have been proven to handle long-range dependencies theoretically and empirically~\cite{gu2021combining} with linear scaling in sequence length, but were computationally prohibitive until Structured State Space Sequence Models (S4)~\cite{gu2021efficiently}. S4 and related architectures by \citet{fu2022hungry,smith2022simplified,poli2023hyena} are based on a new parameterization that relies on time-invariance of the SSM parameters to enable efficient computation. 
Recently, Mamba~\cite{gu2023mamba} improved on S4-based architectures by relaxing the time-invariance constraint on SSM parameters, while maintaining computational efficiency. This allows Mamba to achieve high performance on a range of benchmarks from audio and genomics to language modeling, while maintaining linear scaling in sequence length. In this paper, we benchmark the performance of SSMs like S4 and Mamba on sensory CSP tasks, and show that they consistently outperform LSTMs and Transformers.

\section{\benchmark{}: A Continuous Sequence Prediction Benchmark}
\begin{figure*}[!tbp]
    \centering
   
        \includegraphics[width=\linewidth]{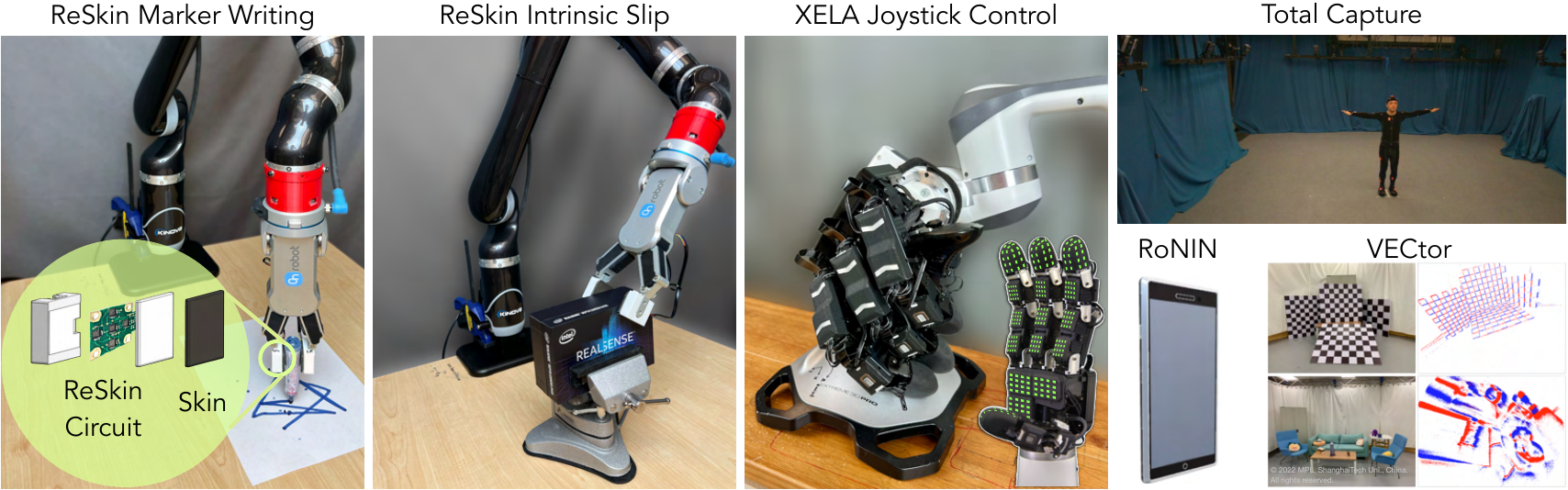}

    \caption{\textbf{\benchmark{}} is comprised of six datasets. Three datasets -- ReSkin Marker Writing, ReSkin Intrinsic Slip and XELA Joystick Control are tactile datasets collected in-house on two different robot setups as demonstrated above. Three other datasets -- RoNIN~\cite{herath2020ronin}, VECtor~\cite{gao2022vector} and TotalCapture~\cite{trumble2017total} are curated open-source datasets.}
    \label{fig:datasets}
    
\end{figure*}

\label{sec:datasets}
We address the scarcity of datasets with dense, continuous labels for sequence-to-sequence prediction by collecting three touch datasets with 1000 trajectories each and combining them with three IMU datasets from literature to create \benchmark{}. For each dataset, we design tasks to predict labeled sequences from \textit{single} sensor data to avoid confounding factors. We also include data from varied sources like cameras and robot movements to facilitate future research in multi-sensor integration and multimodal learning. The detailed characteristics of these datasets are summarized in Table \ref{tab:datasets}, aiming to support diverse sensory data analysis.


\subsection{Touch Datasets}
\label{sec:touch-datasets}

Our touch datasets are collected on two magnetic tactile sensor designs: ReSkin~\cite{bhirangi2021reskin} and Xela~\cite{tomo2018new}. The ReSkin setup consists of a 6-DOF Kinova JACO Gen1 robot with a 1-DOF RG2 OnRobot gripper as shown in Figure \ref{fig:datasets}. Both gripper surfaces are sensorized with a $ 32$ mm $\times$ $30$ mm $\times$ $2$ mm ReSkin sensor. Each sensor has five 3-axis magnetometers which measure changes in magnetic flux resulting from the deformation of the skin on the gripper surface. Appendix \ref{app:reskin-fab} contains more details on the fabrication and integration of ReSkin into the gripper.

The Xela setup consists of a 7-DOF Franka Emika robot fitted with a 16-DOF Allegro hand by Wonik Robotics. Each finger on the hand is sensorized with three 4x4 uSkin tactile sensors and one curved uSkin tactile sensor from XELA Robotics as shown in Figure \ref{fig:datasets}. Sensor integration was provided by XELA robotics, which was designed specifically for the Allegro Hand. While the underlying sensory mode is the same for both ReSkin and Xela, they differ in spatial and temporal resolution, physical layout, and magnetic source. 

\begin{table*}[!t]
    \caption{Summary of all the modalities present in \benchmark{}. Modalities used for training are \textit{italicized}. In addition to the data used for training models, we also release synchronized video and robot kinematics data to facilitate further research in CSP problems.}
    \label{tab:datasets}
    \vskip 0.15in
    \centering
    \begin{tabular}{p{3cm}p{6cm}lp{3cm}l}
    \toprule
        Dataset & Modalities & Model Inputs & Model Outputs & Size \\
         & & (dim) & (dim) & (min) \\\midrule\midrule
        Marker Writing & \textit{ReSkin} (100 Hz), 2 Cameras (30 Hz), \textit{Robot}~(45 Hz) & ReSkin (30) & End-effector \newline velocity (2) & 420 \\\midrule
        Intrinsic Slip & \textit{ReSkin} (100 Hz), 3 Cameras (30 Hz), \textit{Robot}~(45 Hz) & ReSkin (30) & End-effector \newline velocity (3) & 640\\\midrule
        Joystick Control & \textit{Xela} (100 Hz), 2 Cameras (30 Hz), Robot (50 Hz), Hand (300 Hz), \textit{Joystick} (20 Hz) & Xela (552) & Joystick State (3) & 580\\\midrule
        
        VECtor \newline ~\cite{gao2022vector} & \textit{IMU} (200 Hz), 2 Cameras (30 Hz), RGBD (30 Hz), Lidar (10 Hz), \textit{MoCap} (120 Hz)  & IMU (7) & User velocity (3)& 22 \\\midrule
        TotalCapture \newline ~\cite{trumble2017total} & \textit{IMU} (60 Hz), 8 Cameras (60 Hz), \newline \textit{MoCap} (60 Hz) & IMU (39) & Joint velocities (60) & 45 \\\midrule
        RoNIN \newline ~\cite{herath2020ronin} & \textit{IMU} (200 Hz), \textit{3D Tracking Phone} (200 Hz) & IMU (7) & User velocity (2) & 600\\\bottomrule
    \end{tabular}
\end{table*}

\subsubsection{ReSkin: Marker Writing Dataset}
We collect 1000 Kinova robot trajectories of randomized linear strokes across a paper. Initially, the marker is arbitrarily placed between the gripper tips, and data collection begins when the marker touches the paper. The robot then moves linearly between 8-12 random points uniformly sampled within a 10cm x 10cm workspace, pausing for a randomly sampled delay of 1-4 seconds after each motion. Images of sample trajectories can be found in Appendix \ref{app:setup}.

The goal of this sequential prediction problem is to use tactile signal from the gripper to predict the velocity of the end-effector in the plane of the table. Velocity labels are easily obtained from robot kinematics, and serve as a proxy for the velocity of the marker strokes against the paper. What makes this problem challenging is that the sensor picks up contact information from both, the relative motion between the marker and the gripper, and the motion of the marker against the paper. The model must learn to disentangle these two motions to make accurate predictions. 




\subsubsection{ReSkin: Intrinsic Slip Dataset}
\label{sec: boxslip}
We again use the Kinova setup to collect 1000 trajectories of intrinsic slip -- the gripper grasping and slipping along different boxes clamped to a table. At the start of every episode, we close the gripper at a random location and orientation on the box and start recording data. We sample 8-12 random locations and orientations within the workspace of the robot along the length of the box, and then command the robot to move along the box while slipping against it. We use 10 boxes of different sizes to collect this dataset to improve data diversity in terms of contact dynamics. Example images and dimensions are available in Appendix \ref{app:box-slip}.

The goal of the sequential prediction problem is to use the sequence of tactile signals from the gripper tips to predict the translational and rotational velocity of the end-effector (again obtained from robot kinematics) in the plane of the robot's motion. In addition, the abrasive nature of the task causes the skin to wear out over time. To account for this wear, we change the gripper tips and skins after 25 trajectories on every box, improving data diversity as a result.


\subsubsection{XELA: Joystick Control Dataset}
For our final dataset, we record 1000 trajectories of data from the Allegro hand interacting with the joystick as shown in Figure \ref{fig:datasets}. The hand/robot setup is teleoperated using a VR-based system derived from HoloDex~\cite{arunachalam2023holo}. Joystick interactions are logged synchronously with robot data, tactile sensing data, and the camera feed. Specifically, this includes the full robot kinematics (7 DOF Arm at 50 Hz + 16 DOF Hand at 300 Hz), XELA tactile output (552 dim at 100 Hz), and 2 Realsense D435 cameras (1080p at 30 Hz). Each trajectory consists of 25-40 seconds of interaction with the joystick.

The goal of the sequential prediction problem is to use tactile signal from the Xela-sensorized robot hand to predict the state of the joystick, which is recorded synchronously with all the other modalities. The extra challenge with this dataset, in addition to the significantly higher dimensionality of the observation space, is the noisier trajectories resulting from human demos instead of a scripted policy. 

\subsection{Curated Public Datasets}
\label{sec:curated-datasets}
In addition to the tactile datasets we release with this paper, we also test our findings on data from other datasets, particularly ones using IMU sensor data (illustrated in Figure \ref{fig:datasets}) -- the RoNIN dataset~\cite{herath2020ronin} which contains smartphone IMU data from 100 human subjects with ground-truth 3D trajectories under natural human motions, the VECtor dataset~\cite{gao2022vector} -- a SLAM dataset collected across three different platforms, and the TotalCapture dataset -- a 3D human pose estimation dataset.

\section{Hierarchical State-Space Models (\method{})}
\label{sec:approach}
\begin{figure*}[htbp]
    \centering
    \includegraphics[width=.95\linewidth]{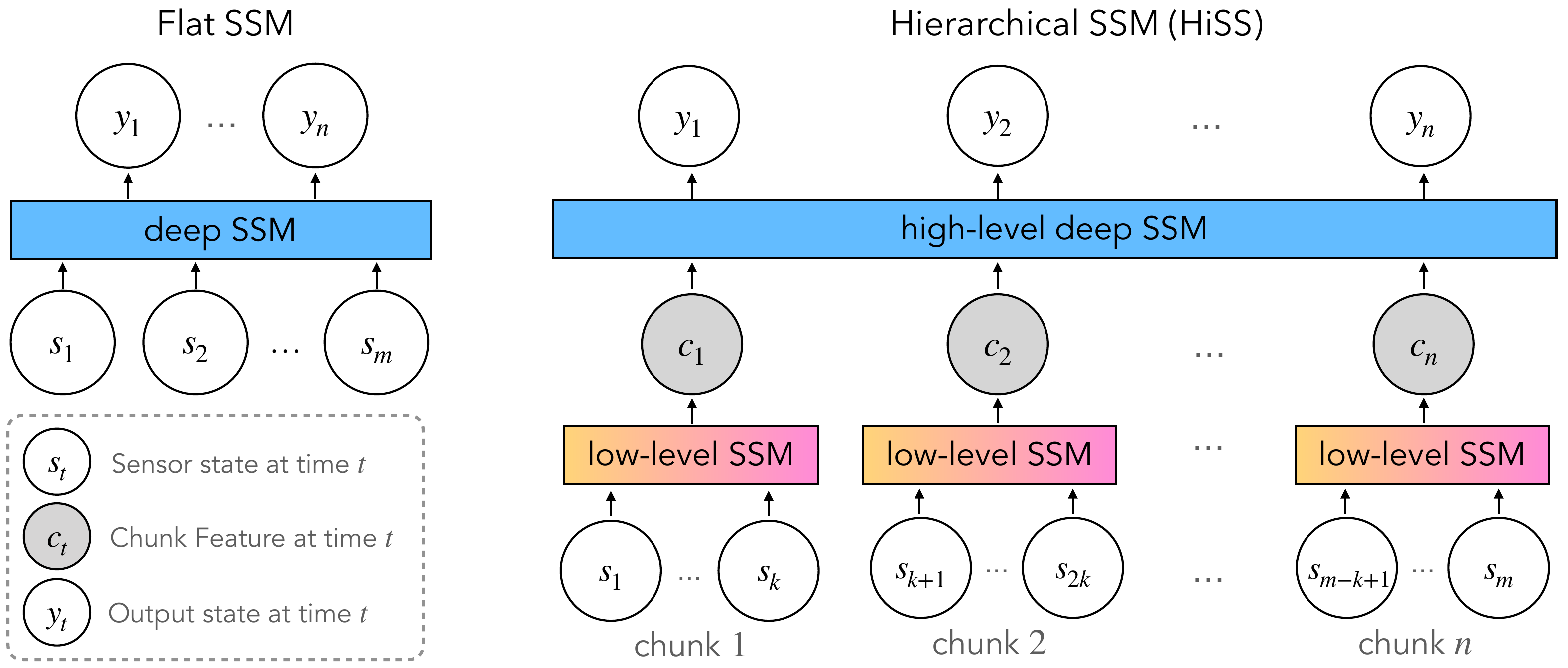}
    \caption{(Left) Flat SSM directly maps a sensor sequence to an output sequence. (Right) \method{} divides an input sequence into chunks which are processed into \textit{chunk features} by a low-level SSM. A high-level SSM maps the resulting sequence to an output sequence.}
    \label{fig:h-v-nh}
\end{figure*}

In this work, we focus on continuous sequence-to-sequence prediction problems for sensors i.e. problems that involve mapping a \textit{sequence} of sensory data to a \textit{sequence} of outputs. In the following sections, we describe our preprocessing pipeline and \method{} -- our approach to sequence-to-sequence reasoning at different temporal scales.

\subsection{Data Preparation and Sampling}
\label{subsec:data-prep}
Every sensor in the real world operates at a different frequency, and data from different sensors is therefore collected at different nominal frequencies. Generally, our sensor sequences come from an inexpensive, noisy sensor operating at a higher frequency than an expensive, high precision device which gives us output sequences. To emulate this scenario and standardize our experiments, all sensor sequences are resampled at a frequency of 50Hz, and output sequences are resampled at 5Hz for all the datasets under consideration, unless specified otherwise. The specific choice of these frequencies is dictated by the sampling frequencies of sensors in the available data. Sequence length ranges (in number of sensor sequence timesteps) for each task are variable: 450-3000 for Marker Writing, 750-2150 for Intrinsic Slip, 750-4250 for Joystick Control, 12000 for the RoNIN, 1900-9100 for VectorEnv and 1700-5600 for TotalCapture. 


All the sensors considered in \benchmark{} are prone to drift; therefore, in line with previous work~\cite{bhirangi2021reskin, guzey2023dexterity, herath2020ronin}, we estimate a resting signal at the start of every sensor trajectory and deviations from this resting signal are passed to the model. Since sensor drift can be causally data-dependent, the entire sensory trajectory is passed to the model as input. Sensor and output sequences are normalized based on data statistics for their corresponding datasets, and details are listed in Appendix \ref{app:model-arch}. Additionally, we find that appending one-step differences to every element in the sensor sequence helps improve performance, in line with numerous prior works~\cite{chen2016s1, holden2016deep}.

\subsection{Model Architecture}
\label{sec:model-arch}



Here we describe Hierarchical State Space Models (\method{}) -- a simple hierarchical architecture that uses SSMs to explicitly reason over sequential data at two temporal resolutions, as shown in Figure \ref{fig:h-v-nh}. The sensor sequence is first divided into a set of equally-sized chunks of size $k$. Each chunk is passed through a shared SSM, say S4, which we refer to as the \textit{low-level} SSM. The outputs of the low-level SSM corresponding to the $k$-th element of each chunk are then concatenated to form a rarified \textit{chunk feature} sequence. Finally, this sequence is passed through a \textit{high-level} sequence model to generate the output sequence.


\textbf{Why should \method{} work?} 
Sequential sensory data is subject to phenomena that occur at different natural frequencies. For instance, an IMU device mounted on a quadrotor is subject to high-frequency vibration noise and low-frequency drift characteristic of MEMS devices~\cite{koksal2018adaptive}. With \method{}, our goal is to create a neural architecture with explicit structure to operate at different temporal scales. This will allow the low-level model to learn effective, temporally local representations, while enabling the high-level model to focus on global predictions over a shorter sequence.

\textbf{Computational Complexity} \method{} builds on top of models like S4 and Mamba which are linear in sequence length, $O(N)$. For non-overlapping chunks of size $k$ each, the low-level model operates on $N/k$ chunks with each computation being $O(k)$. The high-level model in turn operates on a sequence of length $N/k$ resulting in a computation cost $O(N/k)$. The net cost is therefore, $O(k * (N/k) + N/k) = O(N + N/k) = O(N)$. For the case of overlapping chunks, in the most extreme case where we have an overlap of $(k-1)$ elements between chunks, we now have $N$ chunks of size $k$, each operated on by the low-level model. The high-level model operates on the resulting chunk feature sequence of length $N$. Therefore the computational complexity in this case is $O(Nk + N) = O(Nk)$, still significantly better than transformers which have a complexity of $O(N^2)$.

\subsection{Training details}
We focus on sequence-to-sequence prediction tasks. All our models are trained end-to-end to minimize MSE loss as explained in Section \ref{sec:seq-to-seq}. For all tactile datasets and VECtor, we use an 80-20 train-validation split. For the RoNIN dataset, we use the first four minutes of every trajectory for our analysis,  and use a validation set consisting of trajectories from unseen subjects. For TotalCapture, we use the train-validation split proposed by \citet{trumble2017total}. Hyperparameter sweep ranges for each of our models and baselines, along with the resulting range of parameter counts are listed in Appendix \ref{app:model-arch}. We maintain similar ranges of parameter counts across models for the same task.

\begin{table*}[!t]
    \caption{Comparison of MSE prediction losses for baseline and \method{} models on \benchmark{}. Reported numbers are averaged over 5 seeds for the best performing models. MW: Marker Writing, IS: Intrinsic Slip, R: RoNIN, V: VECtor, JC: Joystick Control, TC: TotalCapture}
    \label{tab:main-results}
    \vskip 0.15in
    \centering
    \begin{tabular}{p{2cm}llcccccc}
    \toprule
    Model type & \multicolumn{2}{l}{Model Architecture}& MW & IS & JC & R & V & TC \\
    & & & (cm/s) & & & (m/s) & (m/s) & (m/s)  \\\midrule
    \multirow{6}{*}{Flat} & Transformer &  & 2.3750 & 0.4600 & 1.0200 & - & 0.0432 & - \\ 
    &                       LSTM    &  & 1.1685 & 0.3099 & 1.0740 & 0.0444 & 0.0353 & \textbf{0.1767} \\ 
        &                   S4      &  & 1.3190 & 0.2617 & 0.9804 & 0.0382 & 0.0341 & 0.3483 \\ 
        &                   Mamba   &  & 0.8830 & 0.1757 & 1.0640 & 0.0401 & 0.0319 & 0.3645 \\ 
        &                   MEGA    &  & 0.8960 & 0.2105 & 0.9806 & 0.0370 & 0.0330 & 0.1944 \\
\midrule\midrule & High-level & Low-level & & & & &  \\ 
        \cmidrule{2-3}\multirow{17}{*}{Hierarchical} & \multirow{5}{*}{Transformer} 
                            & Transformer     & 0.6680	& 0.2192	& 0.9112	& 0.0620	& 0.0372 & 0.3048\\
        &                   & LSTM     & 0.9958 & 0.2527 & 0.9350 & 0.0421 & 0.0377 & 0.3197 \\ 
        &                   & S4        & \underline{0.6205} & 0.1574 & \textbf{0.8980} & 0.0363 & 0.0374 & 0.3583 \\ 
        &                   & Mamba    & 1.0268 & 0.2022 & 0.9060 & 0.0472 & 0.0372 & 0.4560 \\ 
        \multicolumn{2}{r}{(MEGA-chunk)}                    & MEGA & 1.1270 & 0.2090 & 1.0450 & 0.0512 & 0.0403 & \underline{0.1940} \\
        \cmidrule{2-9} & \multirow{4}{*}{LSTM} 
                            & Transformer   & 0.7620	& 0.9373	& 1.6090	& 0.3875	& 0.0302	& 0.2943 \\ 
        &                   & LSTM   & 0.8662 & 0.2837 & 1.0760 & 0.0436 & 0.0288 & 0.2522 \\ 
        &                   & S4         & 0.6370 & \underline{0.1526} & 0.9080 & 0.0481 & 0.0322 & 0.3505 \\ 
        &                   & Mamba      & 0.7915 & 0.1925 & 1.0610 & 0.0442 & 0.0286 & 0.3638 \\ 
        \cmidrule{2-9} & \multirow{4}{*}{S4} 
                            & Transformer  & 0.7570	& 0.2898	& 0.9248	& 0.0439	& 0.0295	& 0.2452 \\ 
        &                   & LSTM  & 0.8590 & 0.1805 & 0.9520 & 0.0319 & 0.0293 & 0.2452 \\ 
        &                   & S4 \mycc         & 0.6255\mycc & 0.1551\mycc & 0.9060\mycc & \textbf{0.0265}\mycc & 0.0303\mycc & 0.3438\mycc \\ 
        &                   & Mamba \mycc      & 0.8257\mycc & 0.1823\mycc & 0.9200\mycc & 0.0322\mycc & 0.0294\mycc & 0.4078\mycc \\ 
        \cmidrule{2-9} & \multirow{4}{*}{Mamba} 
                            & Transformer  & 0.7020	& 0.3011	& 0.9553	& 0.0371	& 0.0293	& 0.2064 \\ 
        &                   & LSTM  & 0.7592 & 0.1746 & 0.9640 & 0.0346 & \underline{0.0267} & 0.2428 \\ 
        &                   & S4 \mycc     & \textbf{0.5663}\mycc & \textbf{0.1316}\mycc & \underline{0.9010}\mycc & \underline{0.0302}\mycc & 0.0298\mycc & 0.2527\mycc \\ 
        &                   & Mamba \mycc  & 0.7248\mycc & 0.1678\mycc & 0.9050\mycc & 0.0325\mycc & \textbf{0.0251}\mycc & 0.3762\mycc \\ \midrule
        \multicolumn{3}{l}{{\setlength{\fboxsep}{2pt}\colorbox{lightgray}{HiSS}} improvement over best Flat} & +35.87\% & +25.10\% & +8.10\% & +30.74\% & +21.30\% & -37.36\% \\ \bottomrule
    \end{tabular}
\end{table*}
\section{Experiments and Results}
\label{sec:experiments}

In this section, we evaluate the performance of \method{} models on \csp{} tasks and understand their strengths and limitations. Unless otherwise specified, we use non-overlapping chunks of size 10, and aim to answer the following questions:
\begin{itemize}[leftmargin=*]
    \item How do SSMs compare with LSTMs and Transformers on \benchmark{}?
    \item Can \method{} provide benefits over temporally flat models?
    \item How does chunk size affect the performance of \method{}?
    \item Is \method{} compatible with existing preprocessing techniques like filtering?
    \item How does \method{} perform in low-data regimes?
\end{itemize}

\paragraph{Baselines:}
We use two categories of baselines: Flat and Hierarchical. Flat models consist of LSTMs, Causal Transformers, S4 and Mamba, in addition to MEGA~\cite{ma2022mega}. Hierarchical baselines include variations of \method{} models where the high-level and/or low-level SSMs are replaced by causal transformers and LSTMs, and MEGA-chunk~\cite{ma2022mega}, which is loosely classified as a high-level transformer with a low-level MEGA model. Table \ref{tab:main-results} presents a performance comparison on CSP-Bench for each of these baselines and proposed \method{} models.


\subsection{Performance of Flat models on \benchmark{}}
\label{sec:ssm-v-flat}
At the outset, we see that SSMs -- Mamba and S4, consistently outperform the best-performing Transformer and LSTM models by 10\% and 14\% median MSE respectively across \benchmark{} tasks. The only anomaly is the TotalCapture dataset where the LSTM outperforms all other models. We analyze this later in Section \ref{sec:total-capture}.

\subsection{Improving CSP Performance with \method{}}

\method{} models perform better than the best-performing flat models, SSM or otherwise, with a \textit{further} improvement of $\sim$23\% median MSE across tasks. Among hierarchical models, \method{} models continue to do as well as or better than the others with a relative improvement of $\sim$ 9.8\% median MSE. Further, we make two key observations within models that use a specific high-level architecture: (1) these models consistently outperform corresponding flat models, indicating that temporal hierarchies are effective at distilling information from continuous sensory data; (2) the best models use S4 as the low-level model, indicating that S4 is particularly adept at capturing low-level temporal structure in the data. 

These observations raise a natural question: What is happening under the hood? In the next four sections, we attempt to better understand the working of \method{}.

\begin{table}[tb]
    \caption{Performance comparison with (a) downsampled inputs, (b) low pass filter on input sequences, and (c) fewer training samples}
    \label{tab:ablations}
    \vskip 0.15in
    \centering
    \setlength\tabcolsep{5.5pt}
    \begin{tabular}{lcccccc}
        \toprule
         & MW & IS & JC & R & V & TC \\\midrule
         \multicolumn{6}{l}{\textit{Downsampled inputs}} \\
         Trnsfrmr  & 2.41 & 0.33 & .957 & .116 & .039 & 0.34 \\ 
        LSTM  & 1.92 & 0.27 & .975 & .094 & .034 & \textbf{0.20} \\ 
        S4  & 2.22 & 0.29 & .974 & .081 & .036 & 0.31 \\ 
        Mamba  & 1.96 & 0.26 & .980 & .077 & .033 & 0.25 \\  
        \method{} & \textbf{0.57} & \textbf{0.13} & \textbf{.901} & \textbf{.027} & \textbf{.025} & 0.26 \\ \midrule
         \multicolumn{6}{l}{\textit{Low Pass Filtering}} \\ 
         Trnsfrmr  & 1.79 & 0.31 & 1.01 & - & .034 & 0.38 \\ 
        LSTM  & 1.15 & 0.26 & 1.08 & .038 & .024 & \textbf{0.12} \\ 
        S4  & 1.19 & 0.22 & 0.94 & .031 & .022 & 0.25 \\ 
        Mamba  & 0.78 & 0.14 & 0.95 & \textbf{.030} & \textbf{.018} & 0.17 \\ 
        \method{}  & \textbf{0.55} & \textbf{0.11} & \textbf{0.87} & .036 & .020 & 0.13 \\ 
        \midrule
         \multicolumn{6}{l}{\textit{Smaller Training Dataset}} \\
         Fraction & 0.3 & 0.3 & 0.3 & 0.3 & 0.5 & 0.5 \\ \noalign{\vskip 1mm}
         Trnsfrmr  & 4.30 & 0.85 & 1.237 & - & .046 & 0.54 \\ 
        LSTM  & 1.83 & 0.54 & 1.313 & .053 & .039 & 0.39 \\ 
        S4  & 2.31 & 0.45 & 1.197 & .043 & .038 & 0.43 \\ 
        Mamba  & 1.74 & 0.37 & 1.195 & .039 & .036 & 0.48 \\ \midrule
        \method{}  & \textbf{1.26} & \textbf{0.29} & \textbf{1.106} & \textbf{.034} & \textbf{.029} & \textbf{0.37} \\\bottomrule
    \end{tabular}
\end{table}

\subsection{Does \method{} Simply do Better Downsampling?}
\label{exp:downsampling}
The first question we seek to answer is whether simply downsampling the sensor sequence to the same frequency as the output would do just as well as \method{}. As we see in Table \ref{tab:ablations}, while some flat models with downsampled sensor sequences indeed improve performance over flat models in Table \ref{tab:main-results}, they remain far behind \method{} models. This reinforces our hypothesis that \method{} models distill more information from the sensor sequence than naive downsampling.

One advantage of using hierarchical models is memory efficiency. They can significantly reduce computational load for models like transformers which scale quadratically in the length of the sequence. Using an SSM such as S4 or Mamba as the low-level model can significantly reduce the computational load $\left( O(N^2) \rightarrow O(N^2/k^2)\right)$ for $k \ll N$, where $k$ and $n$ are chunk size and sequence length respectively. Table \ref{tab:main-results} shows that such a model consistently improves performance over a flat causal Transformer across tasks.

\subsection{Effect of Chunk Size on Performance}
\label{exp:chunk-size}
Having established the effectiveness of \method{} relative to conventional sequence modeling architectures, we seek to investigate the effect of a key parameter -- the chunk size -- on the performance of \method{} models. Downsampling the sensor sequences at the output frequency, as presented in Section \ref{exp:downsampling} essentially corresponds to using a chunk size of 1. The rest of the analysis presented so far uses a chunk size of 10, corresponding to the largest non-overlapping chunks that cover the entire sensory sequence given sensor and output sequence frequencies of 50 Hz and 5 Hz respectively. In this section, we conduct two additional experiments with chunk sizes of 5 and 15 and present the results in Table \ref{tab:chunk-size}. We see that while performance improves drastically as the chunk size increases, it plateaus once the chunk size reaches the ratio of the sensor and output frequencies (10 in our case). This behavior can be explained by the fact that chunk sizes smaller than this ratio result in the model never seeing parts of the sensor sequence, while chunk sizes larger than this ratio result in an overlap between chunks. 

\begin{table}[tb]
    \caption{Effect of chunk size on perfomance of \method{} models}
    \label{tab:chunk-size}
    \vskip 0.15in
    \centering
    \begin{tabular}{ccccccc}
         \toprule
         \multicolumn{1}{c}{\multirow{2}{2.5em}{Chunk size}} & \multirow{2}{*}{MW} & \multirow{2}{*}{IS} & \multirow{2}{*}{JC} & \multirow{2}{*}{R} & \multirow{2}{*}{V} & \multirow{2}{*}{TC} \\
         & & & & & &\\\midrule
         5 & 1.18 & 0.20 & .933 & .046 & .033 & 0.32 \\
         10 & 0.57 & 0.13 & .901 & \textbf{.027} & \textbf{.025} & 0.25 \\
         15 & \textbf{0.54} & \textbf{0.12} & \textbf{.899} & .035 & .026 & \textbf{0.24}
 \\\bottomrule
    \end{tabular}
\end{table}

\subsection{Effect of Sensory Preprocessing on Performance}
\label{sec:lowpass}
A common approach to preprocessing noisy sensor data is to design low-pass filters to process the signal before it's passed through the model. To analyze the compatibility of \method{} models with such existing preprocessing techniques, we separately apply order 5 Butterworth filters with 3 different cut-off frequencies to the sensor sequence and report model corresponding to the best cut-off frequency in Table \ref{tab:ablations}. We make two key observations: (1) with the exception of the \method{} model for RoNIN, low pass filtering improves performance across the board; (2) \method{} models still perform comparably with or better than flat models.

With respect to (1), we see that the best-performing \method{} model from Table \ref{tab:main-results} continues to outperform the best flat model using filtered data, implying that the low-pass filter may have filtered useful information could have been used to improve task performance. This points to an important pitfall of handcrafted preprocessing techniques -- they can often filter out information that could have been exploited by a sufficiently potent model. Consequently, the ability of \method{} models to require little to no preprocessing of the input sequence bolsters their credentials to serve as general purpose models for CSP data.



\subsection{How Does \method{} Perform on Smaller Datasets?}
\label{sec:small-dsets}
The lack of a comprehensive benchmark for continuous sequence prediction so far speaks to the difficulty of collecting large, labeled datasets of sensory data. Therefore, performance in low-data regimes could be critical to wider applicability of different sequence modeling architectures. To benchmark this performance, we compare the performance of flat as well as \method{} models on subsets of the training data. While TotalCapture and VECtor are substantially smaller than the other datasets (see Table \ref{tab:datasets}), we include them in this analysis while using a larger fraction of training data than other datasets. Results are presented in Table \ref{tab:ablations}. We only present the best performing \method{} model here for conciseness. The full table can be found in Appendix \ref{app:ablations}.

We see that on smaller fractions of the training dataset, \method{} outperforms flat baselines on \textit{every} task in \benchmark{}. This indicates an important property of \method{} models -- data efficiency. Low-level models operate identically on all of the chunks in the data, allowing them to learn more effective representations from small datasets than flat models. 

\subsection{Failure on TotalCapture}
\label{sec:total-capture}
The most visible failure case for the performance of both flat SSMs as well as \method{} models is on the TotalCapture dataset, where the flat LSTM significantly outperforms all other models. We hypothesize that the high dimensionality of the input and output spaces prevents SSMs from learning sufficiently expressive representations that can filter out high frequency data. This is also evidenced by the higher performance of LSTM low-level models across hierarchical architectures for this dataset, which correlates with the correspondingly higher effectiveness of the flat LSTM over flat SSMs. Further evidence of the inability of SSMs to filter out noise can be found in Section \ref{sec:lowpass}, where the performance of \method{} models nearly matches the LSTM when the input sequence is passed through a lowpass filter. This indicates that the \method{} model struggles to learn the filtering behavior from data here, unlike other datasets where performance remains fairly consistent with and without the lowpass filter.


\label{sec:freq}

\section{Conclusion and Limitations}
We present \benchmark{}, the first publicly available benchmark for Continuous Sequence Prediction, and show that SSMs do better than LSTMs and Transformers on CSP tasks. Then, we propose \method{}, a hierarchical sequence modeling architecture that is more performative, data efficient and minimizes preprocessing needs for CSP problems. However, sequence-to-sequence prediction from sensory data continues to be an open, relatively underexplored problem, and our work indicates significant room for improvement. Moreover, while SSMs show significant promise for CSP tasks, they are relatively new architectures whose strengths and weaknesses are far from being well-understood. Section \ref{sec:total-capture} explains some of the challenges of SSMs, and as a result, \method{}, on high-dimensional prediction problems with small datasets of noisy sensor data. In terms of ease of training, current \method{} models introduce an additional hyperparameter of chunk size. While we provide a preliminary analysis of the effect of chunk size in Section \ref{exp:chunk-size}, optimizing chunk size is an exciting future direction. Finally, CSP-Bench is large, but the number of sensors that can benefit from learned models is larger. We are committed to supporting \benchmark{} and adding more, larger datasets in the future.




\section*{Acknowledgements}
NYU authors are supported by grants from Honda, and ONR award numbers N00014-21-1-2404 and N00014-21- 1-2758. LP is supported by the Packard Fellowship. We also thank Aadhithya Iyer, Gaoyue Zhou, Irmak Guzey, Ulyana Piterbarg, Vani Sundaram and all other members of GRAIL, NYU for their valuable help and feedback throughout this project.

\section*{Impact Statement}
This paper presents work whose goal is to advance the field of Machine Learning. There are many potential societal consequences of our work, none which we feel must be specifically highlighted here.

\bibliography{references}

\begin{thebibliography}{63}
\providecommand{\natexlab}[1]{#1}
\providecommand{\url}[1]{\texttt{#1}}
\expandafter\ifx\csname urlstyle\endcsname\relax
  \providecommand{\doi}[1]{doi: #1}\else
  \providecommand{\doi}{doi: \begingroup \urlstyle{rm}\Url}\fi

\bibitem[Amini et~al.(2011)Amini, Sarrafzadeh, Vahdatpour, and Xu]{amini2011accelerometer}
Amini, N., Sarrafzadeh, M., Vahdatpour, A., and Xu, W.
\newblock Accelerometer-based on-body sensor localization for health and medical monitoring applications.
\newblock \emph{Pervasive and mobile computing}, 7\penalty0 (6):\penalty0 746--760, 2011.

\bibitem[Arunachalam et~al.(2023)Arunachalam, G{\"u}zey, Chintala, and Pinto]{arunachalam2023holo}
Arunachalam, S.~P., G{\"u}zey, I., Chintala, S., and Pinto, L.
\newblock Holo-dex: Teaching dexterity with immersive mixed reality.
\newblock In \emph{2023 IEEE International Conference on Robotics and Automation (ICRA)}, pp.\  5962--5969. IEEE, 2023.

\bibitem[Bhirangi et~al.(2021)Bhirangi, Hellebrekers, Majidi, and Gupta]{bhirangi2021reskin}
Bhirangi, R., Hellebrekers, T., Majidi, C., and Gupta, A.
\newblock Reskin: versatile, replaceable, lasting tactile skins.
\newblock \emph{arXiv preprint arXiv:2111.00071}, 2021.

\bibitem[Bhirangi et~al.(2023)Bhirangi, DeFranco, Adkins, Majidi, Gupta, Hellebrekers, and Kumar]{bhirangi2023all}
Bhirangi, R., DeFranco, A., Adkins, J., Majidi, C., Gupta, A., Hellebrekers, T., and Kumar, V.
\newblock All the feels: A dexterous hand with large-area tactile sensing.
\newblock \emph{IEEE Robotics and Automation Letters}, 2023.

\bibitem[Calandra et~al.(2018)Calandra, Owens, Jayaraman, Lin, Yuan, Malik, Adelson, and Levine]{calandra2018more}
Calandra, R., Owens, A., Jayaraman, D., Lin, J., Yuan, W., Malik, J., Adelson, E.~H., and Levine, S.
\newblock More than a feeling: Learning to grasp and regrasp using vision and touch.
\newblock \emph{IEEE Robotics and Automation Letters}, 3\penalty0 (4):\penalty0 3300--3307, 2018.

\bibitem[Chang et~al.(2019)Chang, Lambert, Sangkloy, Singh, Bak, Hartnett, Wang, Carr, Lucey, Ramanan, et~al.]{chang2019argoverse}
Chang, M.-F., Lambert, J., Sangkloy, P., Singh, J., Bak, S., Hartnett, A., Wang, D., Carr, P., Lucey, S., Ramanan, D., et~al.
\newblock Argoverse: 3d tracking and forecasting with rich maps.
\newblock In \emph{Proceedings of the IEEE/CVF conference on computer vision and pattern recognition}, pp.\  8748--8757, 2019.

\bibitem[Chavarriaga et~al.(2013)Chavarriaga, Sagha, Calatroni, Digumarti, Tr{\"o}ster, Mill{\'a}n, and Roggen]{chavarriaga2013opportunity}
Chavarriaga, R., Sagha, H., Calatroni, A., Digumarti, S.~T., Tr{\"o}ster, G., Mill{\'a}n, J. d.~R., and Roggen, D.
\newblock The opportunity challenge: A benchmark database for on-body sensor-based activity recognition.
\newblock \emph{Pattern Recognition Letters}, 34\penalty0 (15):\penalty0 2033--2042, 2013.

\bibitem[Chen et~al.(2018)Chen, Zhao, Lu, Wang, Markham, and Trigoni]{chen2018oxiod}
Chen, C., Zhao, P., Lu, C.~X., Wang, W., Markham, A., and Trigoni, N.
\newblock Oxiod: The dataset for deep inertial odometry.
\newblock \emph{arXiv preprint arXiv:1809.07491}, 2018.

\bibitem[Chen et~al.(2021)Chen, Zhang, Yao, Guo, Yu, and Liu]{chen2021deep}
Chen, K., Zhang, D., Yao, L., Guo, B., Yu, Z., and Liu, Y.
\newblock Deep learning for sensor-based human activity recognition: Overview, challenges, and opportunities.
\newblock \emph{ACM Computing Surveys (CSUR)}, 54\penalty0 (4):\penalty0 1--40, 2021.

\bibitem[Chen et~al.(2016)Chen, Yang, Ho, Tsai, Chen, Chang, Lai, Wang, Tsao, and Wu]{chen2016s1}
Chen, T.-E., Yang, S.-I., Ho, L.-T., Tsai, K.-H., Chen, Y.-H., Chang, Y.-F., Lai, Y.-H., Wang, S.-S., Tsao, Y., and Wu, C.-C.
\newblock S1 and s2 heart sound recognition using deep neural networks.
\newblock \emph{IEEE Transactions on Biomedical Engineering}, 64\penalty0 (2):\penalty0 372--380, 2016.

\bibitem[Daum(2005)]{daum2005nonlinear}
Daum, F.
\newblock Nonlinear filters: beyond the kalman filter.
\newblock \emph{IEEE Aerospace and Electronic Systems Magazine}, 20\penalty0 (8):\penalty0 57--69, 2005.

\bibitem[Deng et~al.(2009)Deng, Dong, Socher, Li, Li, and Fei-Fei]{deng2009imagenet}
Deng, J., Dong, W., Socher, R., Li, L.-J., Li, K., and Fei-Fei, L.
\newblock Imagenet: A large-scale hierarchical image database.
\newblock In \emph{2009 IEEE conference on computer vision and pattern recognition}, pp.\  248--255. Ieee, 2009.

\bibitem[Devlin et~al.(2018)Devlin, Chang, Lee, and Toutanova]{devlin2018bert}
Devlin, J., Chang, M.-W., Lee, K., and Toutanova, K.
\newblock Bert: Pre-training of deep bidirectional transformers for language understanding.
\newblock \emph{arXiv preprint arXiv:1810.04805}, 2018.

\bibitem[Fu et~al.(2022)Fu, Dao, Saab, Thomas, Rudra, and R{\'e}]{fu2022hungry}
Fu, D.~Y., Dao, T., Saab, K.~K., Thomas, A.~W., Rudra, A., and R{\'e}, C.
\newblock Hungry hungry hippos: Towards language modeling with state space models.
\newblock \emph{arXiv preprint arXiv:2212.14052}, 2022.

\bibitem[Funabashi et~al.(2019)Funabashi, Yan, Geier, Schmitz, Ogata, and Sugano]{funabashi2019morphology}
Funabashi, S., Yan, G., Geier, A., Schmitz, A., Ogata, T., and Sugano, S.
\newblock Morphology-specific convolutional neural networks for tactile object recognition with a multi-fingered hand.
\newblock In \emph{2019 International Conference on Robotics and Automation (ICRA)}, pp.\  57--63. IEEE, 2019.

\bibitem[Gao et~al.(2022)Gao, Liang, Yang, Wu, Wang, Chen, and Kneip]{gao2022vector}
Gao, L., Liang, Y., Yang, J., Wu, S., Wang, C., Chen, J., and Kneip, L.
\newblock Vector: A versatile event-centric benchmark for multi-sensor slam.
\newblock \emph{IEEE Robotics and Automation Letters}, 7\penalty0 (3):\penalty0 8217--8224, 2022.

\bibitem[Gardiol(2000)]{gardiol2000hierarchical}
Gardiol, N.~H.
\newblock Hierarchical memory-based reinforcement learning.
\newblock In \emph{Neural Information Processing Systems (NIPS)}, volume~13, pp.\  1047--1053. MIT Press, 2000.

\bibitem[Geiger et~al.(2013)Geiger, Lenz, Stiller, and Urtasun]{geiger2013vision}
Geiger, A., Lenz, P., Stiller, C., and Urtasun, R.
\newblock Vision meets robotics: The kitti dataset.
\newblock \emph{The International Journal of Robotics Research}, 32\penalty0 (11):\penalty0 1231--1237, 2013.

\bibitem[Gemmeke et~al.(2017)Gemmeke, Ellis, Freedman, Jansen, Lawrence, Moore, Plakal, and Ritter]{gemmeke2017audio}
Gemmeke, J.~F., Ellis, D.~P., Freedman, D., Jansen, A., Lawrence, W., Moore, R.~C., Plakal, M., and Ritter, M.
\newblock Audio set: An ontology and human-labeled dataset for audio events.
\newblock In \emph{2017 IEEE international conference on acoustics, speech and signal processing (ICASSP)}, pp.\  776--780. IEEE, 2017.

\bibitem[Goel et~al.(2022)Goel, Gu, Donahue, and R{\'e}]{goel2022s}
Goel, K., Gu, A., Donahue, C., and R{\'e}, C.
\newblock It’s raw! audio generation with state-space models.
\newblock In \emph{International Conference on Machine Learning}, pp.\  7616--7633. PMLR, 2022.

\bibitem[Gu \& Dao(2023)Gu and Dao]{gu2023mamba}
Gu, A. and Dao, T.
\newblock Mamba: Linear-time sequence modeling with selective state spaces.
\newblock \emph{arXiv preprint arXiv:2312.00752}, 2023.

\bibitem[Gu et~al.(2021{\natexlab{a}})Gu, Goel, and Re]{gu2021efficiently}
Gu, A., Goel, K., and Re, C.
\newblock Efficiently modeling long sequences with structured state spaces.
\newblock In \emph{International Conference on Learning Representations}, 2021{\natexlab{a}}.

\bibitem[Gu et~al.(2021{\natexlab{b}})Gu, Johnson, Goel, Saab, Dao, Rudra, and R{\'e}]{gu2021combining}
Gu, A., Johnson, I., Goel, K., Saab, K., Dao, T., Rudra, A., and R{\'e}, C.
\newblock Combining recurrent, convolutional, and continuous-time models with linear state space layers.
\newblock \emph{Advances in neural information processing systems}, 34:\penalty0 572--585, 2021{\natexlab{b}}.

\bibitem[Guzey et~al.(2023{\natexlab{a}})Guzey, Dai, Evans, Chintala, and Pinto]{guzey2023see}
Guzey, I., Dai, Y., Evans, B., Chintala, S., and Pinto, L.
\newblock See to touch: Learning tactile dexterity through visual incentives.
\newblock \emph{arXiv preprint arXiv:2309.12300}, 2023{\natexlab{a}}.

\bibitem[Guzey et~al.(2023{\natexlab{b}})Guzey, Evans, Chintala, and Pinto]{guzey2023dexterity}
Guzey, I., Evans, B., Chintala, S., and Pinto, L.
\newblock Dexterity from touch: Self-supervised pre-training of tactile representations with robotic play.
\newblock \emph{arXiv preprint arXiv:2303.12076}, 2023{\natexlab{b}}.

\bibitem[Hasani et~al.(2022)Hasani, Lechner, Amini, Liebenwein, Ray, Tschaikowski, Teschl, and Rus]{hasani2022closed}
Hasani, R., Lechner, M., Amini, A., Liebenwein, L., Ray, A., Tschaikowski, M., Teschl, G., and Rus, D.
\newblock Closed-form continuous-time neural networks.
\newblock \emph{Nature Machine Intelligence}, 4\penalty0 (11):\penalty0 992--1003, 2022.

\bibitem[Herath et~al.(2020)Herath, Yan, and Furukawa]{herath2020ronin}
Herath, S., Yan, H., and Furukawa, Y.
\newblock Ronin: Robust neural inertial navigation in the wild: Benchmark, evaluations, \& new methods.
\newblock In \emph{2020 IEEE International Conference on Robotics and Automation (ICRA)}, pp.\  3146--3152. IEEE, 2020.

\bibitem[Holden et~al.(2016)Holden, Saito, and Komura]{holden2016deep}
Holden, D., Saito, J., and Komura, T.
\newblock A deep learning framework for character motion synthesis and editing.
\newblock \emph{ACM Transactions on Graphics (TOG)}, 35\penalty0 (4):\penalty0 1--11, 2016.

\bibitem[Koksal et~al.(2018)Koksal, Jalalmaab, and Fidan]{koksal2018adaptive}
Koksal, N., Jalalmaab, M., and Fidan, B.
\newblock Adaptive linear quadratic attitude tracking control of a quadrotor uav based on imu sensor data fusion.
\newblock \emph{Sensors}, 19\penalty0 (1):\penalty0 46, 2018.

\bibitem[Kong et~al.(2016)Kong, Zhan, Shin, Denny, and Zhang]{kong2016recognizing}
Kong, B., Zhan, Y., Shin, M., Denny, T., and Zhang, S.
\newblock Recognizing end-diastole and end-systole frames via deep temporal regression network.
\newblock In \emph{Medical Image Computing and Computer-Assisted Intervention-MICCAI 2016: 19th International Conference, Athens, Greece, October 17-21, 2016, Proceedings, Part III 19}, pp.\  264--272. Springer, 2016.

\bibitem[Kuchaiev \& Ginsburg(2017)Kuchaiev and Ginsburg]{kuchaiev2017factorization}
Kuchaiev, O. and Ginsburg, B.
\newblock Factorization tricks for lstm networks.
\newblock \emph{arXiv preprint arXiv:1703.10722}, 2017.

\bibitem[Kulkarni et~al.(2016)Kulkarni, Narasimhan, Saeedi, and Tenenbaum]{kulkarni2016hierarchical}
Kulkarni, T.~D., Narasimhan, K., Saeedi, A., and Tenenbaum, J.
\newblock Hierarchical deep reinforcement learning: Integrating temporal abstraction and intrinsic motivation.
\newblock \emph{Advances in neural information processing systems}, 29, 2016.

\bibitem[Lambeta et~al.(2020)Lambeta, Chou, Tian, Yang, Maloon, Most, Stroud, Santos, Byagowi, Kammerer, et~al.]{lambeta2020digit}
Lambeta, M., Chou, P.-W., Tian, S., Yang, B., Maloon, B., Most, V.~R., Stroud, D., Santos, R., Byagowi, A., Kammerer, G., et~al.
\newblock Digit: A novel design for a low-cost compact high-resolution tactile sensor with application to in-hand manipulation.
\newblock \emph{IEEE Robotics and Automation Letters}, 5\penalty0 (3):\penalty0 3838--3845, 2020.

\bibitem[Liu et~al.(2020{\natexlab{a}})Liu, Caruso, Ilg, Dong, Mourikis, Daniilidis, Kumar, Engel, Valada, and Asfour]{TILO}
Liu, W., Caruso, D., Ilg, E., Dong, J., Mourikis, A., Daniilidis, K., Kumar, V., Engel, J., Valada, A., and Asfour, T.
\newblock Tlio: Tight learned inertial odometry.
\newblock \emph{IEEE Robotics and Automation Letters}, PP:\penalty0 1--1, 07 2020{\natexlab{a}}.
\newblock \doi{10.1109/LRA.2020.3007421}.

\bibitem[Liu et~al.(2020{\natexlab{b}})Liu, Caruso, Ilg, Dong, Mourikis, Daniilidis, Kumar, and Engel]{liu2020tlio}
Liu, W., Caruso, D., Ilg, E., Dong, J., Mourikis, A.~I., Daniilidis, K., Kumar, V., and Engel, J.
\newblock Tlio: Tight learned inertial odometry.
\newblock \emph{IEEE Robotics and Automation Letters}, 5\penalty0 (4):\penalty0 5653--5660, 2020{\natexlab{b}}.

\bibitem[Ma et~al.(2022)Ma, Zhou, Kong, He, Gui, Neubig, May, and Zettlemoyer]{ma2022mega}
Ma, X., Zhou, C., Kong, X., He, J., Gui, L., Neubig, G., May, J., and Zettlemoyer, L.
\newblock Mega: moving average equipped gated attention.
\newblock \emph{arXiv preprint arXiv:2209.10655}, 2022.

\bibitem[Maddern et~al.(2017)Maddern, Pascoe, Linegar, and Newman]{maddern20171}
Maddern, W., Pascoe, G., Linegar, C., and Newman, P.
\newblock 1 year, 1000 km: The oxford robotcar dataset.
\newblock \emph{The International Journal of Robotics Research}, 36\penalty0 (1):\penalty0 3--15, 2017.

\bibitem[Mathieu et~al.(2012)Mathieu, Koch, and Callaway]{mathieu2012state}
Mathieu, J.~L., Koch, S., and Callaway, D.~S.
\newblock State estimation and control of electric loads to manage real-time energy imbalance.
\newblock \emph{IEEE Transactions on power systems}, 28\penalty0 (1):\penalty0 430--440, 2012.

\bibitem[Micucci et~al.(2017)Micucci, Mobilio, and Napoletano]{micucci2017unimib}
Micucci, D., Mobilio, M., and Napoletano, P.
\newblock Unimib shar: A dataset for human activity recognition using acceleration data from smartphones.
\newblock \emph{Applied Sciences}, 7\penalty0 (10):\penalty0 1101, 2017.

\bibitem[Moody \& Mark(2001)Moody and Mark]{moody2001impact}
Moody, G.~B. and Mark, R.~G.
\newblock The impact of the mit-bih arrhythmia database.
\newblock \emph{IEEE engineering in medicine and biology magazine}, 20\penalty0 (3):\penalty0 45--50, 2001.

\bibitem[Morrill et~al.(2021)Morrill, Salvi, Kidger, and Foster]{morrill2021neural}
Morrill, J., Salvi, C., Kidger, P., and Foster, J.
\newblock Neural rough differential equations for long time series.
\newblock In \emph{International Conference on Machine Learning}, pp.\  7829--7838. PMLR, 2021.

\bibitem[Orvieto et~al.(2023)Orvieto, Smith, Gu, Fernando, Gulcehre, Pascanu, and De]{orvieto2023resurrecting}
Orvieto, A., Smith, S.~L., Gu, A., Fernando, A., Gulcehre, C., Pascanu, R., and De, S.
\newblock Resurrecting recurrent neural networks for long sequences.
\newblock In \emph{International Conference on Machine Learning}, pp.\  26670--26698. PMLR, 2023.

\bibitem[Pinto et~al.(2016)Pinto, Gandhi, Han, Park, and Gupta]{pinto2016curious}
Pinto, L., Gandhi, D., Han, Y., Park, Y.-L., and Gupta, A.
\newblock The curious robot: Learning visual representations via physical interactions.
\newblock In \emph{Computer Vision--ECCV 2016: 14th European Conference, Amsterdam, The Netherlands, October 11-14, 2016, Proceedings, Part II 14}, pp.\  3--18. Springer, 2016.

\bibitem[Poli et~al.(2023)Poli, Massaroli, Nguyen, Fu, Dao, Baccus, Bengio, Ermon, and R{\'e}]{poli2023hyena}
Poli, M., Massaroli, S., Nguyen, E., Fu, D.~Y., Dao, T., Baccus, S., Bengio, Y., Ermon, S., and R{\'e}, C.
\newblock Hyena hierarchy: Towards larger convolutional language models.
\newblock \emph{arXiv preprint arXiv:2302.10866}, 2023.

\bibitem[Rusch et~al.(2021)Rusch, Mishra, Erichson, and Mahoney]{rusch2021long}
Rusch, T.~K., Mishra, S., Erichson, N.~B., and Mahoney, M.~W.
\newblock Long expressive memory for sequence modeling.
\newblock \emph{arXiv preprint arXiv:2110.04744}, 2021.

\bibitem[Sch{\"u}tze et~al.(2004)Sch{\"u}tze, Campisano, Colas, Schilling, and Vanrolleghem]{schutze2004real}
Sch{\"u}tze, M., Campisano, A., Colas, H., Schilling, W., and Vanrolleghem, P.~A.
\newblock Real time control of urban wastewater systems—where do we stand today?
\newblock \emph{Journal of hydrology}, 299\penalty0 (3-4):\penalty0 335--348, 2004.

\bibitem[Simon(2006)]{simon2006optimal}
Simon, D.
\newblock \emph{Optimal state estimation: Kalman, H infinity, and nonlinear approaches}.
\newblock John Wiley \& Sons, 2006.

\bibitem[Smith et~al.(2022)Smith, Warrington, and Linderman]{smith2022simplified}
Smith, J.~T., Warrington, A., and Linderman, S.~W.
\newblock Simplified state space layers for sequence modeling.
\newblock \emph{arXiv preprint arXiv:2208.04933}, 2022.

\bibitem[Sonar et~al.(2018)Sonar, Yuen, Kramer-Bottiglio, and Paik]{sonar2018any}
Sonar, H.~A., Yuen, M.~C., Kramer-Bottiglio, R., and Paik, J.
\newblock An any-resolution pressure localization scheme using a soft capacitive sensor skin.
\newblock In \emph{2018 IEEE International Conference on Soft Robotics (RoboSoft)}, pp.\  170--175. IEEE, 2018.

\bibitem[Stetco et~al.(2019)Stetco, Dinmohammadi, Zhao, Robu, Flynn, Barnes, Keane, and Nenadic]{stetco2019machine}
Stetco, A., Dinmohammadi, F., Zhao, X., Robu, V., Flynn, D., Barnes, M., Keane, J., and Nenadic, G.
\newblock Machine learning methods for wind turbine condition monitoring: A review.
\newblock \emph{Renewable energy}, 133:\penalty0 620--635, 2019.

\bibitem[Sun et~al.(2020)Sun, Kretzschmar, Dotiwalla, Chouard, Patnaik, Tsui, Guo, Zhou, Chai, Caine, et~al.]{sun2020scalability}
Sun, P., Kretzschmar, H., Dotiwalla, X., Chouard, A., Patnaik, V., Tsui, P., Guo, J., Zhou, Y., Chai, Y., Caine, B., et~al.
\newblock Scalability in perception for autonomous driving: Waymo open dataset.
\newblock In \emph{Proceedings of the IEEE/CVF conference on computer vision and pattern recognition}, pp.\  2446--2454, 2020.

\bibitem[Sutton et~al.(1999)Sutton, Precup, and Singh]{sutton1999between}
Sutton, R.~S., Precup, D., and Singh, S.
\newblock Between mdps and semi-mdps: A framework for temporal abstraction in reinforcement learning.
\newblock \emph{Artificial intelligence}, 112\penalty0 (1-2):\penalty0 181--211, 1999.

\bibitem[Thu \& Han(2021)Thu and Han]{thu2021hihar}
Thu, N. T.~H. and Han, D.~S.
\newblock Hihar: A hierarchical hybrid deep learning architecture for wearable sensor-based human activity recognition.
\newblock \emph{IEEE Access}, 9:\penalty0 145271--145281, 2021.

\bibitem[Tomo et~al.(2018)Tomo, Regoli, Schmitz, Natale, Kristanto, Somlor, Jamone, Metta, and Sugano]{tomo2018new}
Tomo, T.~P., Regoli, M., Schmitz, A., Natale, L., Kristanto, H., Somlor, S., Jamone, L., Metta, G., and Sugano, S.
\newblock A new silicone structure for uskin—a soft, distributed, digital 3-axis skin sensor and its integration on the humanoid robot icub.
\newblock \emph{IEEE Robotics and Automation Letters}, 3\penalty0 (3):\penalty0 2584--2591, 2018.

\bibitem[Trumble et~al.(2017)Trumble, Gilbert, Malleson, Hilton, and Collomosse]{trumble2017total}
Trumble, M., Gilbert, A., Malleson, C., Hilton, A., and Collomosse, J.
\newblock Total capture: 3d human pose estimation fusing video and inertial sensors.
\newblock In \emph{Proceedings of 28th British Machine Vision Conference}, pp.\  1--13, 2017.

\bibitem[Vaswani et~al.(2017)Vaswani, Shazeer, Parmar, Uszkoreit, Jones, Gomez, Kaiser, and Polosukhin]{vaswani2017attention}
Vaswani, A., Shazeer, N., Parmar, N., Uszkoreit, J., Jones, L., Gomez, A.~N., Kaiser, {\L}., and Polosukhin, I.
\newblock Attention is all you need.
\newblock \emph{Advances in neural information processing systems}, 30, 2017.

\bibitem[Wagner et~al.(2020)Wagner, Strodthoff, Bousseljot, Kreiseler, Lunze, Samek, and Schaeffter]{wagner2020ptb}
Wagner, P., Strodthoff, N., Bousseljot, R.-D., Kreiseler, D., Lunze, F.~I., Samek, W., and Schaeffter, T.
\newblock Ptb-xl, a large publicly available electrocardiography dataset.
\newblock \emph{Scientific data}, 7\penalty0 (1):\penalty0 154, 2020.

\bibitem[Wang \& Strodthoff(2023)Wang and Strodthoff]{wang2023s4sleep}
Wang, T. and Strodthoff, N.
\newblock S4sleep: Elucidating the design space of deep-learning-based sleep stage classification models.
\newblock \emph{arXiv preprint arXiv:2310.06715}, 2023.

\bibitem[Warden(2018)]{warden2018speech}
Warden, P.
\newblock Speech commands: A dataset for limited-vocabulary speech recognition.
\newblock \emph{arXiv preprint arXiv:1804.03209}, 2018.

\bibitem[Welch et~al.(1995)Welch, Bishop, et~al.]{welch1995introduction}
Welch, G., Bishop, G., et~al.
\newblock An introduction to the kalman filter.
\newblock 1995.

\bibitem[Yan et~al.(2018)Yan, Shan, and Furukawa]{yan2018ridi}
Yan, H., Shan, Q., and Furukawa, Y.
\newblock Ridi: Robust imu double integration.
\newblock In \emph{Proceedings of the European conference on computer vision (ECCV)}, pp.\  621--636, 2018.

\bibitem[You et~al.(2019)You, Wang, Pal, Eksombatchai, Rosenburg, and Leskovec]{you2019hierarchical}
You, J., Wang, Y., Pal, A., Eksombatchai, P., Rosenburg, C., and Leskovec, J.
\newblock Hierarchical temporal convolutional networks for dynamic recommender systems.
\newblock In \emph{The world wide web conference}, pp.\  2236--2246, 2019.

\bibitem[Yuan et~al.(2017)Yuan, Dong, and Adelson]{yuan2017gelsight}
Yuan, W., Dong, S., and Adelson, E.~H.
\newblock Gelsight: High-resolution robot tactile sensors for estimating geometry and force.
\newblock \emph{Sensors}, 17\penalty0 (12):\penalty0 2762, 2017.

\end{thebibliography}
\bibliographystyle{icml2024}

\newpage
\appendix
\onecolumn
\section{ReSkin fabrication details}
\label{app:reskin-fab}

ReSkin measures the changes in magnetic flux in its X, Y and Z coordinate system, based on the change in relative distance between the embedded magnetic microparticles in an elastomer matrix and a nearby magnetometer. The use of magnetic microparticles enables freedom in regard to the shape and dimensions of the molded skin. In our use case here, we use a skin of thickness 2mm. This section further details the complete fabrication process involved in the sensorized gripper tips we use for the ReSkin setup described in Section \ref{sec:touch-datasets}. Figure \ref{fig:circuitry} illustrates different components of the sensorized gripper.

\subsection{Circuitry}
\begin{figure*}[htbp]
    \centering
    \includegraphics[width=0.35\linewidth]{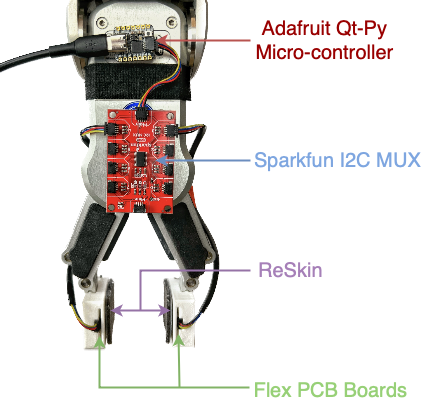}
    \caption{Circuitry}
    \label{fig:circuitry}
\end{figure*}

Data from the ReSkin sensors is streamed to a computer via USB. The two sensors are connected to an I$^2$C MUX which in turn is connected to an Adafruit QT Py microcontroller as described in \citet{bhirangi2021reskin}. See Figure \ref{fig:circuitry}.

\subsection{OnRobot Gripper Tips}
\begin{figure*}[htbp]
    \centering
    \includegraphics[width=0.15\linewidth]{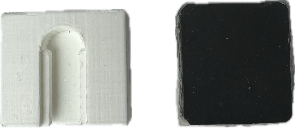}
    \caption{Gripper Tips with ReSkin}
    \label{fig:grippertips}
\end{figure*}

The skins are affixed to the 3D-printed gripper tips using silicone adhesive, as shown in Figure \ref{fig:grippertips}. The dimensions of the tips are $32$ mm $\times 30$ mm $\times 2$ mm. The same tips also house the flex-PCB boards, which measure the change in magnetic flux in all 3 axes. 

\section{Model architectures and Training}
\label{app:model-arch}

\subsection{Flat Architectures}
For each of the flat sequence models presented in this work, the input sequence is first embedded into a hidden state sequence by a linear layer. This hidden state is then passed to the respective sequence model. The outputs of the sequence model (the hidden states for LSTM, S4 and Mamba) are then mapped to the desired output space 

\subsection{Hierarchical architectures}
The hierarchical models are obtained by simply stacking two flat models together. The input sequence is first divided into equal sized chunks as described in Section \ref{sec:model-arch}. Each chunk is passed through the low-level sequence model and the outputs corresponding to the last timestep of each chunk are concatenated to form the chunk feature sequence. This sequence is passed through a high-level sequence model to obtain the output sequence

\subsection{Hyperparameters}
All models are trained for 600 epochs at a constant learning rate of 1e-3. Learning rate schedulers were not found to improve performance by noticeable amounts. Table \ref{tab:flat_hps} contains the ranges of hyperparameters used for training the flat models presented in the paper. We do not sweep over all of these hyperparameters for each task. A subset of these parameters is chosen for each task depending on the input and output dimensionality of the task and the best-performing models are reported.  The exact hyperparameters for each experiment can be found on the Github repository. For any given task, we ensure that sweeps over all model classes consist of models that have the same order of magnitude of learnable parameters.

\begin{table}[htbp]
    \centering
    \begin{tabular}{p{3.5cm}p{3.5cm}p{3.5cm}p{3.5cm}}
        \toprule
         LSTM & {Transformer} & {S4} & {Mamba} \\\midrule
         Input size \newline 16, 32, 64, 128, 256 & Model dim \newline 32, 64, 128, 256, 512 & Model dim \newline 32, 64, 128, 256, 512 & Model dim \newline 32, 64, 128, 256, 512\\ \midrule
         LSTM hidden size \newline	256, 512, 1024 & No. of heads \newline	2,4 \\ \midrule
         No. of layers \newline 2 & No. of layers \newline 4,6 & No. of layers \newline 4, 6 & No. of layers	\newline 4, 6 \\ \midrule
         Dropout \newline 0.0, 0.1 & Dropout \newline 0.0, 0.1 & Dropout \newline 0.0, 0.1 & \\
    \end{tabular}
    \caption{Hyperparameters for flat architectures}
    \label{tab:flat_hps}
\end{table}

For the hierarchical models, we use a smaller subset of the parameters listed in Table \ref{tab:flat_hps} to sweep over the high level models. Parameter ranges swept over for low-level models are listed in Table \ref{tab:hiss_hps}. The exact hyperparameters for each experiment can be found on the Github repository.

\begin{table}[htbp]
    \centering
    \begin{tabular}{p{3.5cm}p{3.5cm}p{3.5cm}}
        \toprule
         LSTM & {S4} & {Mamba} \\\midrule
         Input size \newline 16, 32, 64  & Model dim \newline 16,32,64,128, 256 & Model dim \newline 16, 32, 64, 128, 256 \\ \midrule
         LSTM hidden size \newline	16,32,64,128,256  \\ \midrule
         No. of layers \newline 1 & No. of layers \newline 4, 6 & No. of layers	\newline 3,4 \\ \bottomrule
    \end{tabular}
    \caption{Hyperparameters for low-level models used in hierarchical architectures}
    \label{tab:hiss_hps}
\end{table}

These hyperparameter sweeps result in a range of models with different numbers of parameters. Table \ref{tab:parameter-ranges} lists the range of parameters resulting from the sweeps, and Table \ref{tab:best-parameters} contains the number of parameters in the best-performing models.

\begin{table}[htbp]
    \caption{Range of parameters swept over for baseline and \method{} models on \benchmark{}. Reported numbers are in millions of parameters.}
    \label{tab:parameter-ranges}
    \vskip 0.15in
    \centering
    \begin{tabular}{p{2cm}llcccccc}
    \toprule
    Model type & \multicolumn{2}{l}{Model Architecture}& MW & IS & JC & R & V & TC \\\midrule
    \multirow{4}{*}{Flat} & Transformer &  & 0.4 - 9.5 & 0.4 - 9.5 & 0.7 - 10.6 & - & 0.0 - 0.6 & - \\ 
    &                       LSTM    &  & 3.4 - 13.7 & 0.9 - 3.7 & 3.7 - 14.2 & 0.8 - 3.3 & 0.0 - 0.2 & 3.5 - 13.8 \\ 
        &                   S4      &  & 0.8 - 4.0 & 0.8 - 4.0 & 1.4 - 5.1 & 0.3 - 1.2 & 0.0 - 0.4 & 0.9 - 4.1 \\ 
        &                   Mamba   &  & 0.5 - 10.2 & 1.8 - 10.2 & 0.8 - 11.3 & 0.5 - 2.6 & 0.0 - 0.7 & 0.5 - 10.3\\ 
\midrule\midrule & High-level & Low-level & & & & &  \\ 
        \cmidrule{2-3}\multirow{16}{*}{Hierarchical} & \multirow{4}{*}{Transformer} 
                            & Transformer     & 1.2 - 4.8 & 3.6 - 12.0 & 1.5 - 5.4 & 2.5 - 12.0 & 0.1 - 0.8 & 1.2 - 4.9\\
        &                   & LSTM     & 0.9 - 2.8 & 3.3 - 9.9 & 1.0 - 2.9 & 0.4 - 1.2 & 0.1 - 0.6 & 0.9 - 2.8 \\ 
        &                   & S4        & 1.1 - 3.6 & 3.6 - 10.8 & 1.4 - 4.2 & 0.7 - 2.0 & 0.1 - 0.8 & 1.1 - 3.7 \\ 
        &                   & Mamba    & 1.2 - 4.2 & 3.7 - 11.4 & 1.4 - 4.8 & 0.4 - 2.6 & 0.1 - 0.7 & 1.2 - 4.3 \\ 
        \cmidrule{2-9} & \multirow{4}{*}{LSTM} 
                            & Transformer   & 0.7 - 3.3 & 1.3 - 5.9 & 0.9 - 3.9 & 1.0 - 5.9 & 0.1 - 0.4 & 0.7 - 3.4 \\ 
        &                   & LSTM   & 0.3 - 1.3 & 1.0 - 3.9 & 0.4 - 1.5 & 1.0 - 3.9 & 0.1 - 0.3 & 0.3 - 1.4 \\ 
        &                   & S4         & 0.5 - 2.2 & 1.3 - 4.7 & 0.8 - 2.8 & 1.3 - 4.7 & 0.1 - 0.4 & 0.6 - 2.3 \\ 
        &                   & Mamba      & 0.6 - 2.8 & 1.4 - 5.3 & 0.9 - 3.3 & 0.9 - 5.3 & 0.1 - 0.4 & 0.6 - 2.8 \\ 
        \cmidrule{2-9} & \multirow{4}{*}{S4} 
                            & Transformer  & 0.7 - 3.6 & 1.2 - 6.5 & 1.0 - 4.2 & 0.9 - 6.5 & 0.0 - 0.6 & 0.7 - 3.7 \\ 
        &                   & LSTM  & 0.4 - 1.6 & 0.9 - 4.4 & 0.4 - 1.7 & 0.5 - 1.6 & 0.0 - 0.4 & 0.4 - 1.6 \\ 
        &                   & S4         & 0.6 - 2.5 & 1.2 - 5.3 & 0.8 - 3.0 & 0.8 - 2.4 & 0.1 - 0.6 & 0.6 - 2.5 \\ 
        &                   & Mamba      & 0.6 - 3.0 & 1.3 - 5.9 & 0.9 - 3.6 & 0.5 - 3.0 & 0.1 - 0.5 & 0.7 - 3.1 \\ 
        \cmidrule{2-9} & \multirow{4}{*}{Mamba} 
                            & Transformer  & 0.9 - 5.1 & 0.9 - 5.1 & 1.2 - 5.6 & 0.6 - 5.1 & 0.0 - 0.9 & 0.9 - 5.1 \\ 
        &                   & LSTM  & 0.6 - 3.0 & 0.6 - 3.0 & 0.6 - 3.2 & 0.6 - 3.0 & 0.0 - 0.7 & 0.6 - 3.1 \\ 
        &                   & S4     & 0.8 - 3.9 & 0.9 - 3.9 & 1.0 - 4.5 & 0.9 - 3.9 & 0.1 - 0.9 & 0.8 - 4.0 \\ 
        &                   & Mamba  & 0.8 - 4.5 & 1.0 - 4.5 & 1.1 - 5.0 & 0.5 - 4.5 & 0.1 - 0.8 & 0.9 - 4.5 \\ \bottomrule
    \end{tabular}
\end{table}
\begin{table}[htbp]
    \caption{Parameter count for best-performing baseline and \method{} models on \benchmark{}. Reported numbers are in millions of parameters.}
    \label{tab:best-parameters}
    \vskip 0.15in
    \centering
    \begin{tabular}{p{2cm}llcccccc}
    \toprule
    Model type & \multicolumn{2}{l}{Model Architecture}& MW & IS & JC & R & V & TC \\\midrule
    \multirow{4}{*}{Flat} & Transformer &  & 6.3 & 0.6 & 2.9 & - & 0.4 & - \\ 
    &                       LSTM    &  & 13.7 & 3.7 & 14.2 & 0.9 & 0.2 & 13.8 \\ 
        &                   S4      &  & 4.0 & 4.0 & 5.1 & 0.8 & 0.4 & 0.9 \\ 
        &                   Mamba   &  & 10.2 & 2.6 & 7.9 & 0.7 & 0.7 & 0.5 \\ 
\midrule\midrule & High-level & Low-level & & & & &  \\ 
        \cmidrule{2-3}\multirow{16}{*}{Hierarchical} & \multirow{4}{*}{Transformer} 
                            & Transformer     & 1.2 & 3.6 & 1.5 & 4.0 & 0.2 & 2.5\\
        &                   & LSTM     & 0.9 & 3.6 & 2.9 & 0.6 & 0.4 & 2.5 \\ 
        &                   & S4        & 3.6 & 4.4 & 2.2 & 1.5 & 0.5 & 2.1 \\ 
        &                   & Mamba    & 2.9 & 3.7 & 3.1 & 0.4 & 0.7 & 2.9 \\ 
        \cmidrule{2-9} & \multirow{4}{*}{LSTM} 
                            & Transformer   & 0.7 & 3.7 & 1.7 & 1.0 & 0.2 & 0.9 \\ 
        &                   & LSTM   & 1.3 & 1.3 & 0.5 & 1.1 & 0.1 & 0.6 \\ 
        &                   & S4         & 2.2 & 2.1 & 2.3 & 1.3 & 0.2 & 1.6 \\ 
        &                   & Mamba      & 2.7 & 4.0 & 1.5 & 1.0 & 0.3 & 2.2 \\ 
        \cmidrule{2-9} & \multirow{4}{*}{S4} 
                            & Transformer  & 0.9 & 1.2 & 1.9 & 1.0 & 0.3 & 2.9 \\ 
        &                   & LSTM  & 1.3 & 3.1 & 1.3 & 1.6 & 0.3 & 1.6 \\ 
        &                   & S4         & 2.5 & 4.0 & 2.1 & 0.8 & 0.3 & 2.5 \\ 
        &                   & Mamba      & 3.0 & 5.9 & 3.2 & 0.5 & 0.4 & 0.8 \\ 
        \cmidrule{2-9} & \multirow{4}{*}{Mamba} 
                            & Transformer  & 2.4 & 2.2 & 1.2 & 2.7 & 0.2 & 2.2 \\ 
        &                   & LSTM  & 0.8 & 2.2 & 2.3 & 1.9 & 0.1 & 3.0 \\ 
        &                   & S4     & 3.0 & 3.0 & 3.2 & 1.7 & 0.2 & 3.1 \\ 
        &                   & Mamba  & 4.5 & 2.5 & 3.3 & 0.8 & 0.6 & 2.1 \\ \bottomrule
    \end{tabular}
\end{table}


\clearpage
\newpage \section{Experimental Setup and Data Collection details}
\label{app:setup}

\begin{figure*}[htb]
    \centering
    \includegraphics[width=1\linewidth]{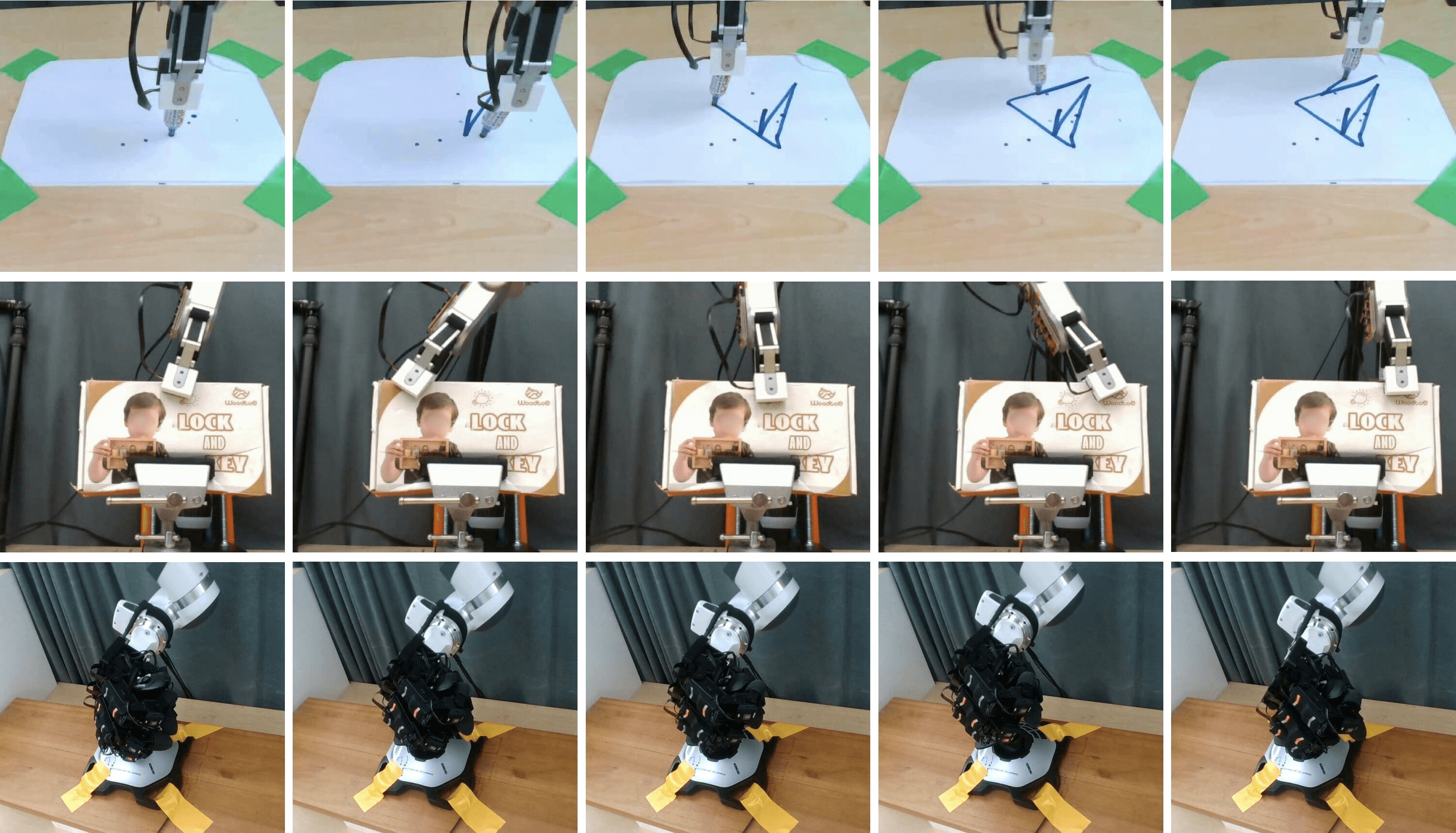}
    \caption{Marker Writing Frames (Top): The gripper tips hold the marker and bring it in contact with the paper before the sequence starts. The arm maneuvers the marker to execute eight strokes on the paper. Instrinsic Slip Frames (Middle): The gripper tips hold the box to start the sequence, and slip through the robot workspace with different orientations. Joystick Control Frames (Bottom): After the sequence begins, the hand holds the joystick, controlling its movement through various positions.}
    \label{fig:mw_frames}
\end{figure*}


\subsection{ReSkin: Onrobot Gripper on a Kinova JACO Arm}
\subsubsection{Marker Writing}
For this experiment, we first grasp the marker with 300 N force in an arbitrary position and bring it in contact with the paper. We then start recording data and command the robot to move sequentially to 8-12 randomly sampled locations within a $10 \times 10 \: cm^2$ plane workspace, making linear strikes on the paper. Figure \ref{fig:mw_frames} illustrates a sample sequence from this dataset. We note that during the strikes, the grasped marker undergoes orientation drifts at times, which adds to the complexity in signal. We record a total of 1000 trajectories of 15-30 seconds each, comprising of 2 different colored markers. The prediction task here is to predict the strike velocity ($\delta$x/$\delta$t, $\delta$y/$\delta$t), given the tactile signals thus reconstructing the overall trajectory.

\begin{figure}
    \centering
    \includegraphics[width=0.4\linewidth]{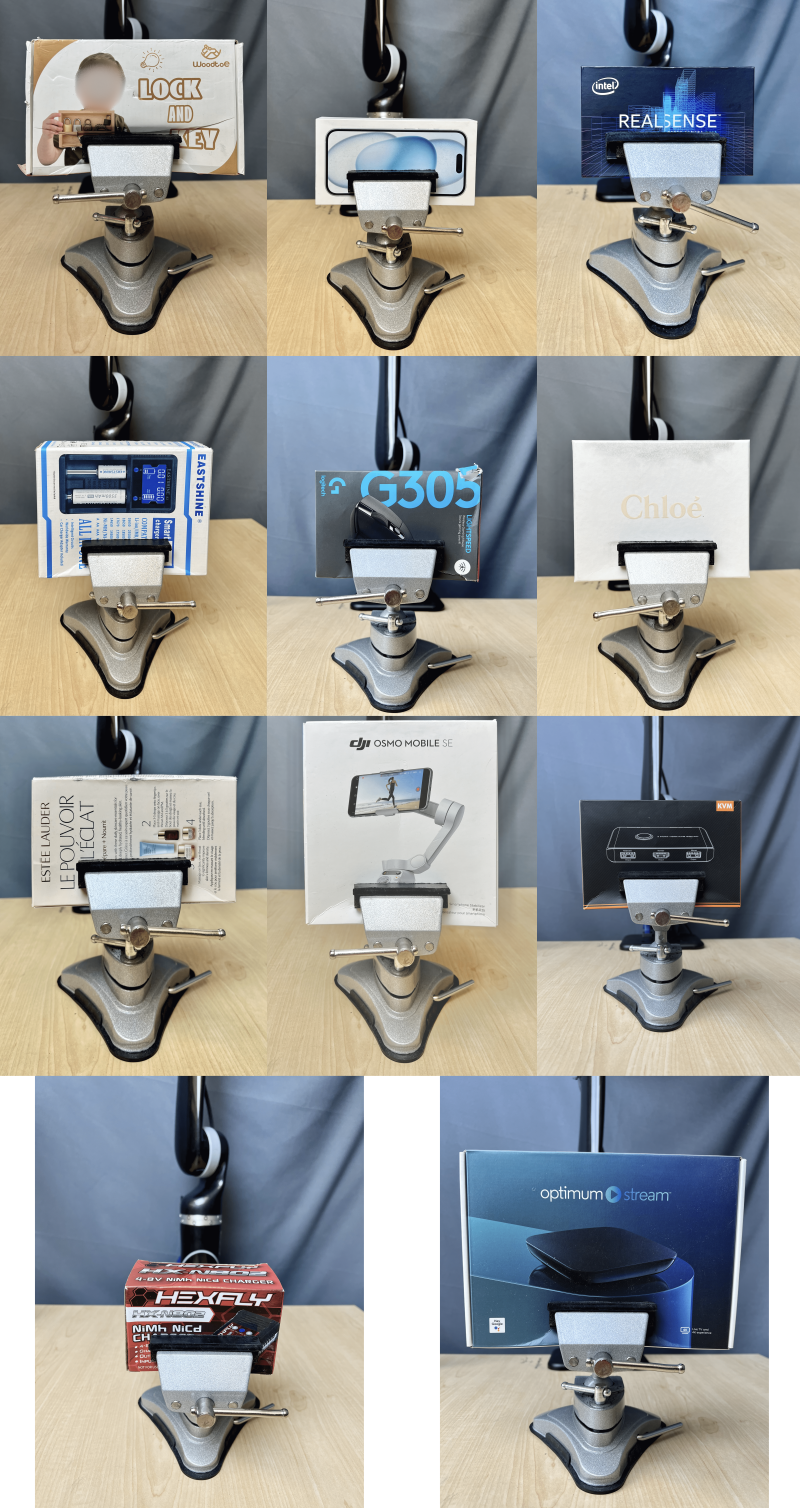}
    \captionof{figure}{Boxes in the Dataset}
    \label{fig:boxes}
\end{figure}

\subsubsection{Intrinsic Slip}
\label{app:box-slip}
\begin{figure}[htbp]
    \centering
    \includegraphics[width=0.3\linewidth]{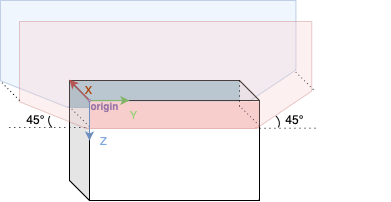}
    \caption{End-effector Workspace on the Box, \& Local Co-ordinate System}
    \label{fig:workspace}
\end{figure}

In Section \ref{sec: boxslip}, we outlined our methodology for collecting data through a total of 1000 trajectories. This involved using 10 distinct boxes and 4 sets of skins for 25 trajectories per combined pair. We first sample a random location and orientation within the task workspace. Next, we close the gripper with a random force sampled in the range of 50-75 N and then start recording data. With the gripper grasping the box, we uniformly sample 8-12 locations sequentially, thus slipping through the robot workspace. Figure \ref{fig:mw_frames} illustrates a sample sequence from this dataset. The workspace is the upper region of the box, which is a space of dimensions \texttt{Box Length x Tip Size(3cm)}, shown in Figure \ref{fig:workspace}. We clamp the wrist rotation limits at [-$\pi$/4, $\pi$/4], making the overall local sampling bounds of the gripper tip position (center of tip), Y:[0, box length], Z:[0, tip size], $\theta$:[-$\pi$/4, $\pi$/4].

\citet{bhirangi2021reskin} characterize the ability of ReSkin sensor models generalize to skins outside the training distribution, but these experiments are limited to single-frame, static data. Here, we collect an analogous dataset for the sequence-to-sequence prediction problem. To avoid confounding effects, the evaluations provided in this paper are based on a random partitioning of this dataset. However, we collect and publish an additional 100 trajectories on an unseen box and an unseen set of skins to test the generalizability of trained models.

The dimensions of all boxes used in this experiment are detailed below. See Table \ref{table:box-dimensions} and Figure \ref{fig:boxes}.

In this experiment, in addition to predicting the linear velocities of the end-effector, we also predict the angular velocities at the wrist/the end-effector rotation ($\delta$x/$\delta$t, $\delta$y/$\delta$t, $\delta$$\theta$/$\delta$t).

\begin{table}[htbp]
    \centering
    \begin{tabular}{|c|c|}
        \hline
        \textbf{Box Number} & \textbf{Dimensions (L x H x W cm)} \\
        \hline
        1 & 20 x 12 x 4 \\
        2 & 16.5 x 8.5 x 3 \\
        3 & 14 x 9 x 5 \\
        4 & 17 x 13 x 4.5 \\
        5 & 15 x 10 x 4.5 \\
        6 & 16.5 x 13 x 6 \\
        7 & 17 x 10 x 5.5 \\
        8 & 18 x 19.5 x 5.5 \\
        9 & 17 x 11 x 3.5 \\
        10 & 12 x 8 x 6.5 \\
        \hline
        11 (unseen) & 23 x 16 x 5 \\
        \hline
    \end{tabular}
    \captionof{table}{Dimensions of Boxes in the Dataset}
    \label{table:box-dimensions}
\end{table}

\subsection{Xela: Allegro Hand on a Franka Emika Panda Arm}

\subsubsection{Joystick Control}

\begin{figure*}[htbp]
    \centering
    \includegraphics[width=0.2\linewidth]{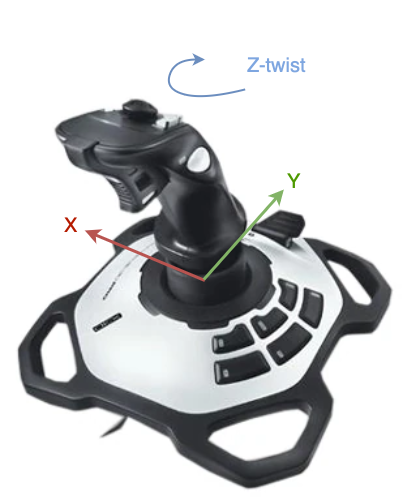}
    \caption{Extreme3D Pro Joystick \& Co-ordinate System}
    \label{fig:dataset_3}
\end{figure*}

For the final tactile dataset, we teleoperate an Allegro Hand with Xela sensors mounted on a Franka arm to interact with an Extreme3D Pro Joystick shown in Figure \ref{fig:dataset_3}, which streams data comprising of 6 rotation axes (X, Y, Rz, Throttle, Hat0X, Hat0Y) and 12 buttons (Trigger, 2 Thumb Buttons, 2 Top Buttons, 1 Pinkie Button and 6 Base Buttons). Unlike the prior datasets, which originated out of random yet scripted policies, this dataset has an added complexity from the unstructured human interactive control. Figure \ref{fig:mw_frames} illustrates a sample sequence from this dataset. Due to the arm workspace and the finger size constraints, we focus on 3 axes - X, Y and Z-twist for our prediction task. Given the readings from the Xela sensors, we predict the joystick's states of interest. 

\newpage
\section{Results and Ablations}
\label{app:ablations}
\subsection{Standard deviations for reported results}
The results presented in Table \ref{tab:main-results} are averaged over 5 random seeds each. Table \ref{tab:std-results} presents the standard deviations over seeds for each of the tasks and models.

\begin{table}[htbp]
    \caption{Comparison of standard deviation in MSE over 5 seeds for baseline and \method{} models on \benchmark{}.}
    \label{tab:std-results}
    \vskip 0.15in
    \centering
    \begin{tabular}{p{2cm}llcccccc}
    \toprule
    Model type & \multicolumn{2}{l}{Model Architecture}& MW & IS & JC & R & V & TC \\
    & & & (cm/s) & & & (m/s) & (m/s) & (m/s)  \\\midrule
    \multirow{4}{*}{Flat} & Transformer &  & 0.0805& 0.0161& 0.0544& - & 0.0004 & - \\ 
    &                       LSTM    &  & 0.0540 & 0.0184 & 0.0006 & 0.0074 & 0.0014 & 0.0039 \\ 
        &                   S4      &  & 0.0634 & 0.0159 & 0.0188 & 0.0049 & 0.0012 & 0.0172 \\ 
        &                   Mamba   &  & 0.0224 & 0.0111 & 0.1060 & 0.0040 & 0.0011 & 0.0064 \\ 
\midrule\midrule & High-level & Low-level & & & & &  \\ 
        \cmidrule{2-3}\multirow{16}{*}{Hierarchical} & \multirow{4}{*}{Transformer} 
                            & Transformer     & 0.0438 & 0.0164 & 0.0250 & 0.0057 & 0.0013 & 0.0159\\
        &                   & LSTM     & 0.0429 & 0.0250 & 0.0420 & 0.0039 & 0.0016 & 0.0114 \\ 
        &                   & S4        & 0.0215 & 0.0084 & 0.0188 & 0.0021 & 0.0028 & 0.0416 \\ 
        &                   & Mamba    & 0.0617 & 0.0145 & 0.0180 & 0.0054 & 0.0015 & 0.0202 \\ 
        \cmidrule{2-9} & \multirow{4}{*}{LSTM} 
                            & Transformer   & 0.0359 & 0.0120 & 0.0721 & 0.1826 & 0.0017 & 0.0257 \\ 
        &                   & LSTM   & 0.0310 & 0.0093 & 0.0244 & 0.0022 & 0.0012 & 0.0121 \\ 
        &                   & S4         & 0.0405 & 0.0069 & 0.0295 & 0.0022 & 0.0014 & 0.0038 \\ 
        &                   & Mamba      & 0.1174 & 0.0179 & 0.0199 & 0.0049 & 0.0014 & 0.0143 \\ 
        \cmidrule{2-9} & \multirow{4}{*}{S4} 
                            & Transformer  & 0.0545 & 0.0273 & 0.0172 & 0.0031 & 0.0015 & 0.0030 \\ 
        &                   & LSTM  & 0.0511 & 0.0099 & 0.0255 & 0.0012 & 0.0014 & 0.0069 \\ 
        &                   & S4         & 0.0274 & 0.0076 & 0.0238 & 0.0009 & 0.0008 & 0.0179 \\ 
        &                   & Mamba      & 0.0357 & 0.0044 & 0.0136 & 0.0024 & 0.0012 & 0.0151 \\ 
        \cmidrule{2-9} & \multirow{4}{*}{Mamba} 
                            & Transformer  & 0.0499 & 0.0154 & 0.0500 & 0.0050 & 0.0007 & 0.0077 \\ 
        &                   & LSTM  & -  & 0.0142 & 0.0131 & 0.0030 & 0.0013 & 0.0171 \\ 
        &                   & S4     & 0.0453 & 0.0066 & 0.0347 & 0.0019 & 0.0016 & 0.0088 \\ 
        &                   & Mamba  & 0.0542 & 0.0042 & 0.0313 & 0.0022 & 0.0010 & 0.0156 \\ \bottomrule
    \end{tabular}
\end{table}

\subsection{Sensor Data Preprocessing with Filtering}
In this section, we provide more detailed tables for the experiments in Sections \ref{sec:lowpass}. Table \ref{tab:app-lowpass} contains results from separately applying order 3 Butterworth filters to the input sequences with cutoff frequencies of 0.75Hz, 2.5Hz and 7.5Hz. For each setting, we pick the set of models corresponding to the cutoff frequency with the best performance, and report average performance over 3 seeds.
\begin{table*}[htbp]
    \caption{Comparison of MSE prediction losses for flat and \method{} models on \benchmark{} when passing the input sequences through a low-pass filter. Reported numbers are averaged over 5 seeds for the best performing models. MW: Marker Writing, IS: Intrinsic Slip, JC: Joystick Control, TC: TotalCapture}
    \label{tab:app-lowpass}
    \vskip 0.15in
    \centering
    \begin{tabular}{p{2cm}llcccccc}
    \toprule
    Model type & \multicolumn{2}{l}{Model Architecture}& MW & BS & JC & RoNIN & VECtor & TC \\
    & & & (cm/s) & & & (m/s) & (m/s) & (m/s)  \\\midrule
    \multirow{4}{*}{Flat} & Transformer &  & 1.7940 & 0.3096 & 1.0080 &  - & 0.0346 & 0.3845 \\ 
    &                       LSTM    &  & 1.1498 & 0.2596 & 1.0770 & 0.0382 & 0.0242 & \textbf{0.1234} \\ 
        &                   S4      &  & 1.1885 & 0.2209 & 0.9449 & \underline{0.0305} & 0.0228 & 0.2467 \\ 
        &                   Mamba   &  & 0.7823 & 0.1367 & 0.9459 & \textbf{0.0297} & \textbf{0.0188} & 0.1661 \\ 
\midrule\midrule & High-level & Low-level & & & & &  \\ 
        \cmidrule{2-3}\multirow{12}{*}{Hierarchical} & \multirow{3}{*}{Transformer} & LSTM     & 1.0052 & 0.1883 & 0.9074 & 0.0532 & 0.0284 & 0.2314 \\ 
        &                   & S4        & 0.6703 & 0.1249 & \textbf{0.8652} & 0.0434 & 0.0260 & 0.2908 \\ 
        &                   & Mamba    & 0.8912 & 0.1251 & 0.8731 & 0.0435 & 0.0243 & 0.3118 \\ 
        \cmidrule{2-9} & \multirow{3}{*}{LSTM} & LSTM   & 0.8063 & 0.2434 & 1.0500 & 0.0430 & 0.0272 & 0.1754 \\ 
        &                   & S4         & \underline{0.6462} & 0.1477 & 0.9885 & 0.0419 & 0.0288 & 0.1968 \\ 
        &                   & Mamba      & 0.7515 & 0.1657 & 1.0080 & 0.0420 & 0.0234 & 0.1755 \\ 
        \cmidrule{2-9} & \multirow{3}{*}{S4} & LSTM  & 0.8525 & 0.1390 & 0.9269 & 0.0306 & 0.0272 & 0.1905 \\ 
        &                   & S4 \mycc         & 0.6667\mycc & 0.1221\mycc & 0.9296\mycc & 0.0377\mycc & 0.0222\mycc & 0.2284\mycc \\ 
        &                   & Mamba \mycc      & 0.7825\mycc & 0.1180\mycc & 0.8898\mycc & 0.0396\mycc & \underline{0.0207}\mycc & 0.2527\mycc \\ 
        \cmidrule{2-9} & \multirow{3}{*}{Mamba} & LSTM  & 0.8143 & 0.1308 & 0.9660 & 0.0369 & 0.0255 & 0.1594 \\ 
        &                   & S4 \mycc     & \textbf{0.5535}\mycc & \underline{0.1074}\mycc & \underline{0.8665}\mycc & 0.0362\mycc & 0.0272\mycc & \underline{0.1301}\mycc \\ 
        &                   & Mamba \mycc  & 1.5657\mycc & \textbf{0.1057}\mycc & 0.8765\mycc & 0.0367\mycc & 0.0212\mycc & 0.1466\mycc \\ \bottomrule
    \end{tabular}
\end{table*}

\subsection{Smaller Datasets}
In this section, we provide more detailed tables for the experiments in Sections \ref{sec:small-dsets}. Table \ref{tab:small-dsets} contains results from subsampling the training datasets -- 30\% of the dataset for MW, IS, JC and RoNIN, and 50\% of the dataset for VECtor and TotalCapture. We see that \method{} consistently outperforms flat models across tasks in \benchmark{} when training on fractions of the training dataset, indicating the sample efficiency of \method{} models.
\begin{table*}[tbp]
    \caption{Comparison of MSE prediction losses for flat and \method{} models on \benchmark{} when using a fraction of the training dataset. Reported numbers are averaged over 5 seeds for the best performing models. MW: Marker Writing, IS: Intrinsic Slip, JC: Joystick Control, TC: TotalCapture}
    \label{tab:small-dsets}
    \vskip 0.15in
    \centering
    \begin{tabular}{p{2cm}llcccccc}
    \toprule
    Model type & \multicolumn{2}{l}{Model Architecture}& MW & IS & JC & RoNIN & VECtor & TC \\
    & & & (cm/s) & & & (m/s) & (m/s) & (m/s)  \\
    & (Fraction) & & 0.3 & 0.3 & 0.3 & 0.3 & 0.5 & 0.5 \\\midrule
    \multirow{4}{*}{Flat} & Transformer &  & 4.2975 & 0.8509 & 1.2370 & - & 0.0460 & 0.5430 \\ 
    &                       LSTM    &  & 1.8322 & 0.5376 & 1.3130 & 0.0533 & 0.0390 & 0.3855 \\ 
        &                   S4      &  & 2.3070 & 0.4450 & 1.1970 & 0.0431 & 0.0379 & 0.4338 \\ 
        &                   Mamba   &  & 1.7443 & 0.3677 & 1.1950 & 0.0394 & 0.0358 & 0.4838 \\ 
\midrule\midrule & High-level & Low-level & & & & &  \\ 
        \cmidrule{2-3}\multirow{6}{*}{Hierarchical} & \multirow{3}{*}{S4} & LSTM  & \underline{1.5417} & 0.3428 & 1.2350 & 0.0387 & 0.0331 & 0.3982 \\ 
        &                   & S4 \mycc         & 1.5460\mycc & \underline{0.2931}\mycc & 1.1260\mycc & \textbf{0.0346}\mycc & 0.0337\mycc & 0.3992\mycc \\ 
        &                   & Mamba \mycc      & 2.3302\mycc & 0.3760\mycc & \textbf{1.1060}\mycc & 0.0412\mycc & 0.0326\mycc & 0.4913\mycc \\ 
        \cmidrule{2-9} & \multirow{3}{*}{Mamba} & LSTM  & 1.5810 & 0.3478 & 1.2410 & \underline{0.0362} & \underline{0.0309} & \textbf{0.3530} \\ 
        &                   & S4 \mycc     & \textbf{1.2600}\mycc & \textbf{0.2883}\mycc & 1.1370\mycc & 0.0378\mycc & 0.0333\mycc & \underline{0.3675}\mycc \\ 
        &                   & Mamba \mycc  & 1.7508\mycc & 0.3688\mycc & \underline{1.1140}\mycc & 0.0383\mycc & \textbf{0.0286}\mycc & 0.4320\mycc \\ \bottomrule
    \end{tabular}
\end{table*}

\newpage
\section{TotalCapture Preprocessing}
This dataset provides readings from 12 IMU sensors and the ground truth poses of 21 joints obtained from the Vicon motion capture system. To standardize the data within a consistent coordinate system, we transformed all IMU sensor readings from their native IMU frames to the Vicon frame. Our task is to predict the velocities of the 21 joints given the IMU acceleration data in the Vicon reference frame.

To convert IMU acceleration data into the Vicon frame, we utilize the calibration results provided in the files named \texttt{<subject\_id>\_<sequence\_name>\_calib\_imu\_ref.txt} and \texttt{<sequence\_name>\_Xsens\_AuxFields.sensors}. The acceleration of each IMU sensor in the Vicon frame is calculated as follows:
\begin{equation}
    a_{\text{vicon}} = R_{\text{inertial}}^{\text{vicon}} R_{\text{imu}}^{\text{inertial}} a_{\text{imu}},
\end{equation}
where $R_{\text{imu}}^{\text{inertial}}$ is the rotation matrix converted from the IMU local orientation quaternion (w, x, y, z) provided in the \texttt{<sequence\_name>\_Xsens\_AuxFields.sensors} files. This quaternion represents the IMU's orientation in the inertial reference frame.

Furthermore, $R_{\text{inertial}}^{\text{vicon}}$ is obtained by converting the quaternion information (\texttt{<imu\_name> x y z w}) available in the \texttt{<subject\_id>\_<sequence\_name>\_calib\_imu\_ref.txt} files, which encapsulates the transformation from the inertial frame to the Vicon global frame.

\end{document}